\documentclass[onecolumn,12pt]{article}
\usepackage[top=.75in, bottom=.75in, left=.75in, right=.75in]{geometry}
\usepackage{amsmath,amsfonts,amscd,amssymb}
\usepackage{graphicx}
\usepackage{epstopdf}
\usepackage{overpic}
\usepackage{cancel}
\usepackage{rotating}
\usepackage{url}
\usepackage{caption}
\usepackage{color}
\usepackage{rotating}
\usepackage{multirow}
\usepackage{wrapfig}
\usepackage{mathtools}
\usepackage{subeqnarray}
\usepackage{bm}
\usepackage{setspace}
\usepackage{palatino} 
\usepackage[numbers,sort&compress]{natbib}

\usepackage[bottom,flushmargin,hang,multiple]{footmisc}
\usepackage{lipsum}
\newcommand\blfootnote[1]{%
  \begingroup
  \renewcommand\thefootnote{}\footnote{#1}%
  \addtocounter{footnote}{-1}%
  \endgroup
}

\definecolor{header1}{cmyk}{0,0,0,1}

\usepackage[utf8]{inputenc}

\usepackage[normalem]{ulem}
\usepackage{color}

\title{\vspace{-.55in}{\fontsize{16}{16}\selectfont \textbf{Improving aircraft performance using machine learning: a review
}}\vspace{-.15in}}

\author{\normalsize{Soledad Le Clainche$^{1}$, Esteban Ferrer$^{1}$, Sam Gibson$^{2}$,}\\ \normalsize{Elisabeth Cross$^{2}$, Alessandro Parente$^{3}$, Ricardo Vinuesa$^{4}$}\\
\footnotesize{$^1$ Universidad Polit\'ecnica de Madrid, Spain} \\
\footnotesize{$^2$ University of Sheffield, United Kingdom}\\
\footnotesize{$^3$ Universit\'e Libre de Bruxelles, Belgium}\\
\footnotesize{$^4$ KTH Royal Institute of Technology, Sweden}
}

\date{}

\begin{document}
\maketitle

\blfootnote{$^*$ Corresponding author: XX}
\vspace{-.2in}
\begin{abstract}
This review covers the new developments in machine learning (ML) that are impacting the multi-disciplinary area of aerospace engineering, including fundamental fluid dynamics (experimental and numerical), aerodynamics, acoustics, combustion and structural health monitoring. 
We review the state of the art, gathering the advantages and challenges of ML methods across different aerospace disciplines and provide our view on future opportunities.  The basic concepts and the most relevant strategies for ML are presented together with the most relevant applications in aerospace engineering, 
revealing that ML is improving aircraft performance and that these techniques will have a large impact in the near future. 
%
\end{abstract}

\tableofcontents
\section{Introduction}\label{sec:intro}

Climate change and increasing resource scarcity are challenges that Europe needs to face in the coming decades. All this has a direct impact on air transport, which is struggling to maintain its performance and competitiveness while ensuring a development focused on sustainable mobility. Research and innovation are essential to maintain the capabilities of the aviation industry, driven by the rise of new markets and new competitors as a result of globalization. A new long-term vision for the aeronautics sector is essential to ensure its successful advancement. In this line, new requirements for the future aviation industry have been defined by the ACARE Flightpath 2050, a Group of Recognized Personalities in the aeronautic sector, including stakeholders from the aeronautics industry, air traffic management, airports, airlines, energy providers and the research community. Aeronautics and air transport comprises both: air vehicle and system technology. The future of aviation should focus on improving design, reducing manufacturing time and cost (including certification and upgrade processes), and also improving the parts forming the overall air travel system (general aviation, aircraft, airlines, airports, air traffic management and maintenance, repair and overhaul). 

The ACARE Flightpath 2050 has defined 5 goals that should be achieved by 2050 to guarantee the path through sustainable mobility:
\begin{itemize}
\item Compared to the capabilities of a typical aircraft in the year 2000, by 2050 new technologies should allow 90\% reduction in NOx emissions, 75\% reduction in CO2 emissions per passenger/kilometre and 65\% reduction in noise emission of flying aircraft.
\item When taxiing, aircraft movements should be emission-free.
\item Novel strategies to design and manufacture aerial vehicles should be developed to make them recyclable.
\item Sustainable alternative fuels should be developed to position Europe as the centre of excellence in the field, and sustained by a strong European energy policy.
\item By 2025, Europe should take the lead to establish global environmental standards, formulating and prioritizing an environmental action plan, and being at the forefront of atmospheric research. 
\end{itemize}
At the same time, ensuring safety and security is also a major priority, with the aim at reducing by 80\% the number of accidents by 2050 compared to 2000, taking into account the rising traffic.

To achieve these goals, it is extremely important to find newer eco-friendly alternatives suitable for the industry to reduce the aviation net carbon emissions and noise. To this aim, the aerospace industry is gathering efforts towards developing new aerodynamic designs, more efficient, reducing the oil consumption whilst maintaining the safety in the flight performance. Moreover, finding new alternatives to fossil fuels, improving the energy efficiency in combustion systems, or finding optimal routes for air traffic management (ATM) are also some of the key points where the aerospace industry should advance to minimize the environmental footprint. However, to achieve these objectives it is not enough to improve the ‘standard’ configurations. The aerospace engineering industry is aware that to go beyond the state-of-the-art it is necessary to develop novel ground-breaking disruptive technologies.

\begin{figure}[h!]
	\centering
	\includegraphics[height = 0.4\textwidth]{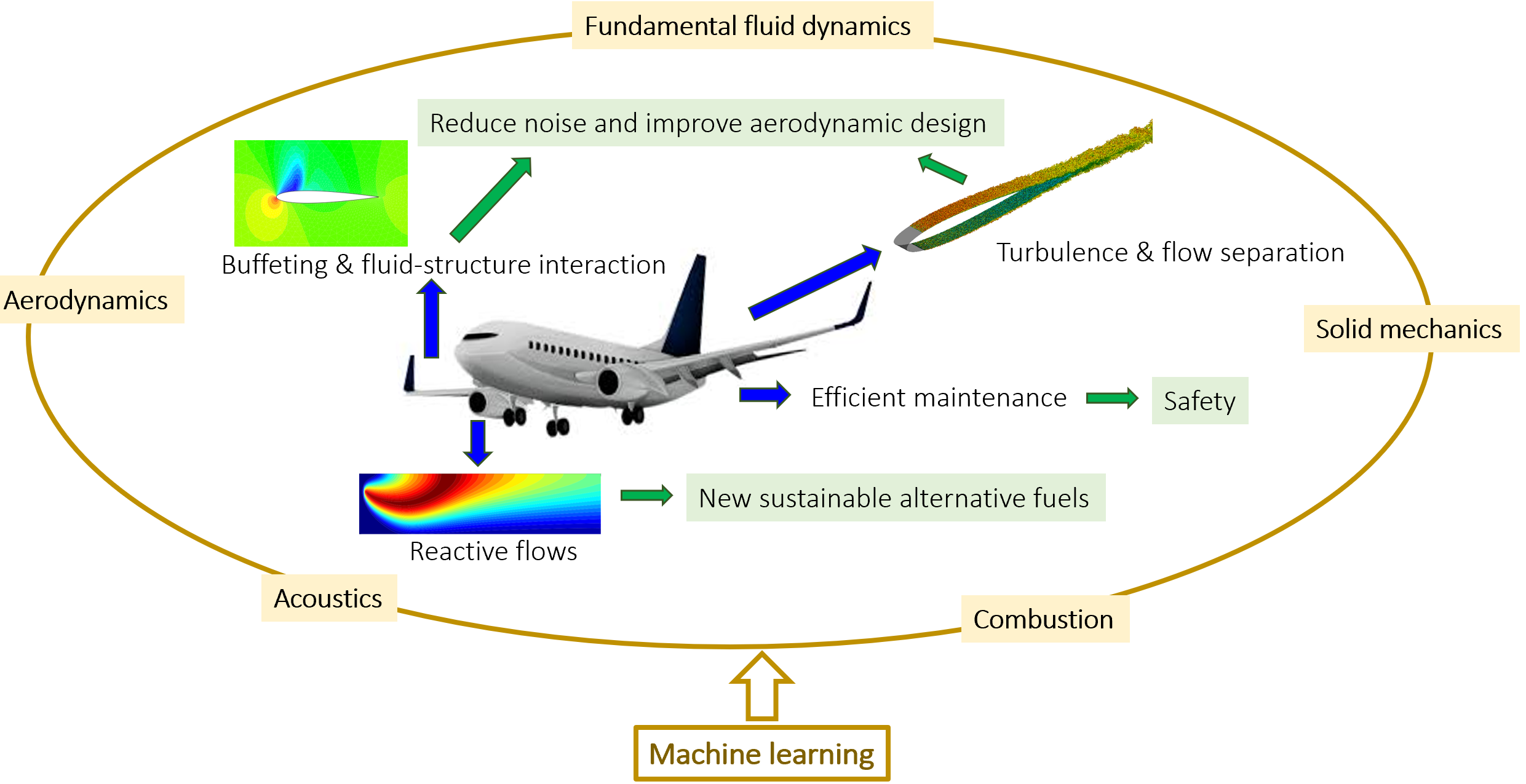}
	\caption{Towards sustainable aviation using machine learning. \label{figINTRO}}
\end{figure}

Fluid and solid mechanics need to be advanced with applications in aerodynamics, acoustics, and combustion, to develop new technologies, resulting in novel aircraft designs with reduced environmental impact (see Fig. \ref{figINTRO}). Researchers in collaboration with the aeronautical industry should explore: (i) new aircraft configurations able to reduce noise and pollution emissions, (ii) cruise drag reduction by manipulating turbulent flow structures close to the aircraft surface ({\it i.e.,} delaying the boundary transition from laminar to turbulent flow), using novel friendly low-risk practices, (iii) novel strategies for flow control (rising the benefits achieved by only changing the external shape) to enhance the aerodynamic performance reducing drag, noise or flow transition, rising lift or controlling unsteadiness or flow separation, and (iv)  reduce the system complexity with novel aircraft materials and lighter designs (which directly results in less fuel consumption), and reduced aircraft maintenance and life cost cycle~\cite{ValeroAST2013}.

High-fidelity numerical simulations and advanced experimental techniques ({\it i.e.,} wind tunnel experiments or open-air experiments as in the case of flight test, et cetera) allow collecting a large variety of data, containing relevant information about physical principles connected to the aerodynamic performance of the aircraft, the efficiency of the combustion system or the main instabilities driving the flow dynamics and the possibility of attenuating or boosting such instabilities using active or passive flow control techniques~\cite{AbadiaHerediaetalESWA2022}. Additionally, experiments ({\it i.e.,} ultrasounds, non-intrusive testing, et cetera) and simulations provide information connected to the fundamentals of solid mechanics, the presence of noise and the structural health of the aircraft, allowing for noise control and early failure detection. 

However, the economic and computational cost, related to the performance of experiments and simulations, encourages researchers to look for new alternatives, which allow to advance in the field {\it i.e.,} developing relevant technologies for the aerospace industry, while avoiding delays in the manufacturing and time-to-market process. Aircraft development, manufacturing, maintenance and support are four critical levels that must be accurate and reliable to ensure the success of the aerospace industry. 

Artificial intelligence (AI) and machine learning (ML) have been introduced in the aerospace industry for various applications connected to the reduction of aircraft's environmental impact, including data interpretation~\cite{LeClaincheetalJFM2022}, system management, customer service or aircraft modelling and to generate new high-fidelity databases at a reduced (economic and CPU) cost~\cite{LopezMartinetal2022}, solving problems of optimization, flow control, or even providing optimal sensors distributions for solid mechanics or aeroelasticity applications. In the recent review article by Brunton {\em et al.}~\cite{BruntonetalMLRev2021}, the authors summarize the new trends and perspective of ML in the aerospace industry, including its application for smart manufacturing, and in the development (and aircraft design), production and product support phases (aircraft design,
manufacturing, verification and validation). The authors reveals the possibilities of ML to process data in light computations increasing the production rate, based on the idea of the use of ML techniques that are measurable, interpretable and certifiable.

ML is generally understood as a branch of AI, although there are nuances in the definition: ML aims at improving systems performance using self-learning algorithms, while AI tries to mimic natural intelligence solving complex problems and enabling decision making (although not maximizing the system efficiency)~\cite{PandeyetalJTURB2020}. 
Both AI and ML are connected to Big Data, a term that linked to the enormous volume of  data that floods the aeronautical sector every day or that is generated from CFD simulations or experimental measurements and connected to aircraft aerodynamic performance~\cite{MendezetalAST2020,CorrochanoetalJAE2022}. Combining Big Data with ML techniques, it is possible to develop reduced dimensional systems, such as reduced order models (ROMs)~\cite{VegaLeClainche2020,LeClainche19}, capable to accurately predict the evolution of the flow dynamics~\cite{AbadiaHerediaetalESWA2022,LeClaincheFerrerEnergies18,doi:10.1063/1.5119342} ({\it i.e.,} flow control, reduce cruise drag  \cite{srinivasan2019predictions}, boundary layer transition, etc.), or surrogate models, capable to predict the aerodynamic forces and moments acting on the aircraft as a function of some parameters ({\it i.e.,} Reynolds number, Mach number,  geometry shape, etc.).

This review provides a state-of-the-art of AI and ML applications in the aerospace engineering field. The basic concepts and most relevant strategies for ML and AI are brought together to explain the similarities found in the nomenclature of similar techniques used in different fields, also shedding light on new applications of these algorithms, quite extended in other fields but not known to the aerospace industry. For example, the review details the use of machine learning for reduced order modelling, which can be used to accelerate numerical simulations, or for temporal and spatial forecast (including non-intrusive sensing). Additionally, we include relevant applications of the field, including flow control, acoustics, combustion, flight test and structural health monitoring. 

This article intends to explore the possibilities of ML, an emerging field for the aerospace industry, identifying new research lines of potential interest and bringing new ideas to developing the technologies of the future, and founded on a primary goal: to fight climate change. This article reviews the main disciplines connected to the aerodynamic performance of the aircraft (see Fig. \ref{figINTRO}): fundamental fluid dynamics, aerodynamics, acoustics, combustion, and general solid mechanics; and based on the idea of finding novel efficient designs, capable to reduce noise and pollutant emissions, while at the same time, ensuring safety and security. 

The article is organized as follows. Section \ref{sec:MLmethod} introduces the machine learning methodologies, and the literature review of the main applications in aerospace engineering is presented in Section \ref{sec:FluidDyn} for fundamental fluid dynamics, Section \ref{sec:Aerodyn} for aerodynamics, Section  \ref{sec:Acourtics} for aeroacoustics, Section \ref{sec:combustion} for combustion and Section \ref{sec:Solid} for solid mechanics. The main conclusions are presented in Section \ref{sec:conclusions}.

\section{Machine learning methodology: a general overview\label{sec:MLmethod}}

\begin{figure}[h!]
	\centering
	\includegraphics[height = 0.35\textwidth]{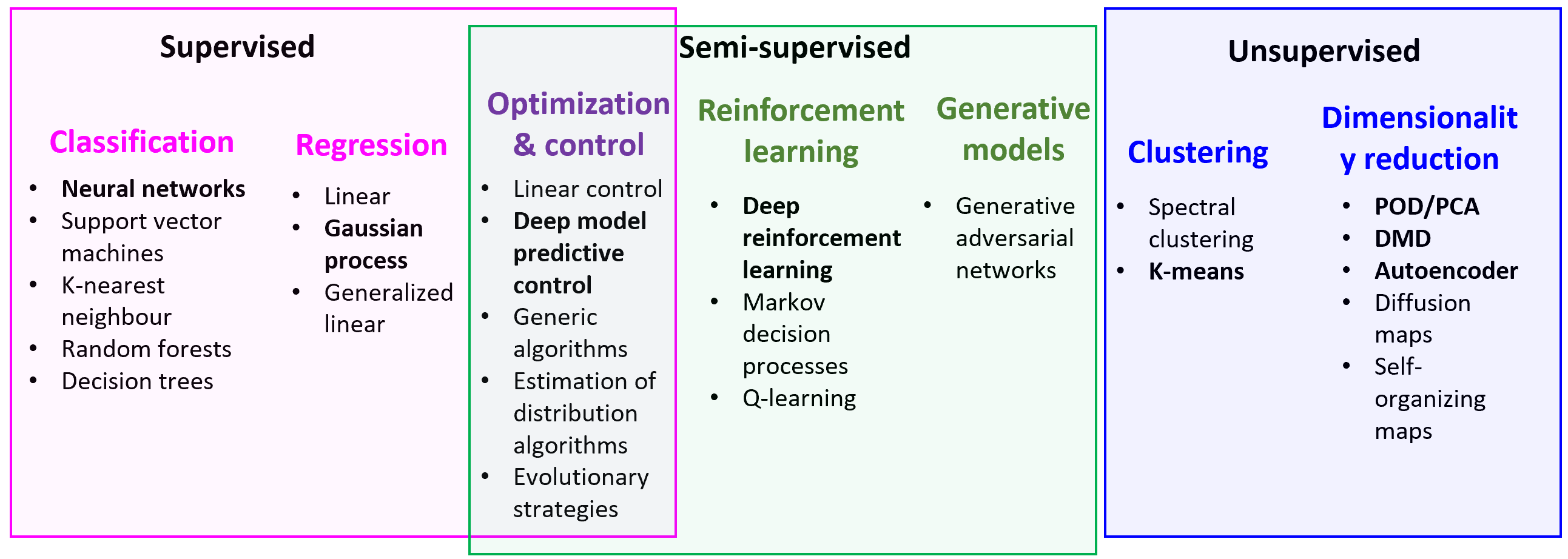}
	\caption{Machine learning methods: a general overview. Classification extracted from Ref. \cite{BruntonetalAARR2019}. In bold, the most popular techniques in the field of aerospace engineering. \label{fig:MLclass}}
\end{figure}

Current advances in computer science are strongly related to the increasing amount of data generated and stored in the different disciplines conforming aerospace engineering. The valuable information contained in these databases encourages researchers to develop and test sophisticated algorithms to exploit such information, to gain insight and knowledge from the data and to subsequently propose and develop new commercial strategies aligned with the ideas behind the concept of sustainable aviation: developing new cleaner and safer aircraft designs. ML is a fast-growing science in the field of aerospace engineering due to its good capabilities to extract information from complex databases, which it later used to develop models, such as ROMs or surrogate models. Based on available data and the type of training carried out within the analysis, ML algorithms can be classified into unsupervised, semi-supervised or supervised learning, as presented in Fig. \ref{fig:MLclass}. This section briefly introduces the basic idea behind some of these ML algorithms, which have been used for different applications in the field of aerospace engineering. 

\subsection{Neural networks} 


ML uses artificial neural networks (ANNs), also called as neural networks (NNs), to process and extract information from databases. The name of this computing system is inspired by the biological neural networks of the human brain. ML uses NNs to solve an optimization problem. More specifically, using back propagation and stochastic gradient descent algorithms, ML optimizes the following compound function 
\begin{equation}
\mathop{\arg \min}\limits_{A_j}  (f_P(\textbf{A}_P,\cdots,f_2(\textbf{A}_2,f_1(\textbf{A}_1,\textbf{X}))\cdots)+\lambda g(\textbf{A}_j)),\label{eq:optimization}
\end{equation}
which represents a NN formed by $P$ layers. The weights connecting years $k$ and $k+1$ are given in matrix $\textbf{A}_k$. To accurately represent the data, which is the main goal of ML and NNs, it is necessary to regularize this system, which is massively undetermined, with the function bias $g(\textbf{A}_j)$. Using different regularization strategies, it is possible to prevent {\em overfitting}, allowing the NNs to generalize the solution obtained from the training to different data sequences.

\begin{figure}[h!]
	\centering
	\includegraphics[height = 0.3\textwidth]{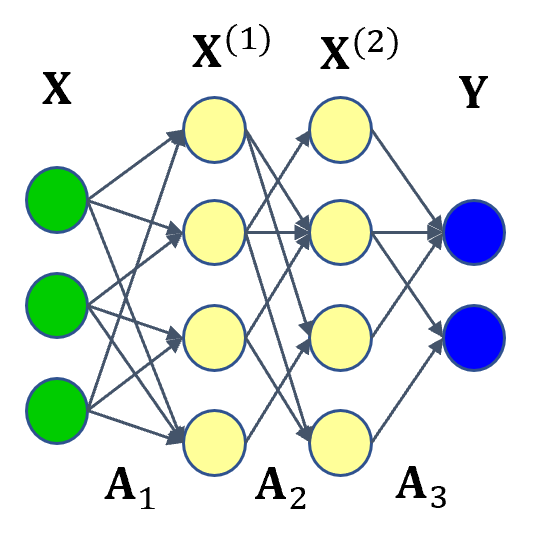}
	\caption{Sketch representing a NN architecture with  three layers. Structure extracted from Ref.~\cite{LeClaincheetalJFM2022}.  \label{fig:NN}}
\end{figure}

To define the architecture of a NN, it is necessary to set the dimension and number of layers and the type of connections and mappings (linear/non-linear) between the different layers, in order to get the best system performance to represent data. Depending on all these parameter and on the way the optimization problem is defined, it is possible to define the different type of ML architectures previously presented in Fig. \ref{fig:MLclass}. 
The generic structure of a multi-layer NN is presented in Fig. \ref{fig:NN}. 

In this example, the input database  $\textbf{X}=[x_1\; x_2\; x_3] \in \mathbb{R}^3$, is map $\textbf{Y}=[y_1\; y_2]$ in the output layer, with a different space dimension, $\mathbb{R}^2$. The relationship between the different layers in this NN is defined in eq. (\ref{eq:NN3nl})  as 
\begin{equation}
\begin{split}
 \textbf{X}^{(1)} & = f_1(\textbf{A}_1, \textbf{X}) \\
 \textbf{X}^{(2)} & = f_2(\textbf{A}_2, \textbf{X}^{(1)}) \\
 \textbf{Y}\; & = f_3(\textbf{A}_3, \textbf{X}^{(2)}). \\ \label{eq:NN3nl}
\end{split}
\end{equation}
This expression is also defined by the following compound functions $f_j$, defined for $P$ layers as
\begin{equation}
\textbf{Y}=f_P(\textbf{A}_P,\cdots,f_2(\textbf{A}_2,f_1(\textbf{A}_1,\textbf{X}))\cdots), \label{eq:NNcompMnl}
\end{equation}
which defines the general optimization problem presented in eq. (\ref{eq:optimization}).  This expression defines a system of equations that is highly under-determined, which requires additional constrains to be solved, where the $P$ matrices generate the best possible mapping.

The activation function $f_j$ (for the layer $j$), can  be both linear or non-linear ({\it i.e.,} sigmoid, $\tanh$, Rectified Linear Unit -ReLU, $\cdots$).
In aerodynamics, radial basis function neural networks (RBFNNs) have been used since the past, due to the good generalization capabilities of the RBF as activation function~\cite{Haykin2009}. This NNs model, proposed by Broomhead \& Lowe~\cite{BroomheadLowe} for the first time, 
generally uses a non-linear activation function $f_j$ defined by a Gaussian basis function as
\begin{equation}
    f_j(\textbf{x}_k)=\exp\left(-\frac{\|\textbf{x}_k-\textbf{v}_j^2 \|}{2\sigma_j^2}\right),\label{eq:RBF}
\end{equation}
where the two hyper-parameters $\textbf{v}$ and $\sigma$ are known as the center vector and the width parameter, respectively. The value set for these two parameter strongly affects the NN performance \cite{KouZhang2017,KouZhang2018}.

Depending on the type of connection between the layers and the neuron of the NNs, it is possible to identify different architectures. For instance, Recurrent Neural Networks (RNNs) process sequence of data using recurrent connections between the different layers. Among the different architectures of RNNs it is possible to distinguish Long Short-Term Memory (LSTM) networks, which in sequence prediction problems are capable of learning order dependence. Another type of architecture widely use for ML applications in fluid dynamics are Convolutional Neural Networks (CNNs). This type of networks transfer information locally between the different neurons and layers. Based on a threshold, each neuron may transfer or not transfer information to the next layer of neurons. More details describing deep learning basic concepts can be found in the book by Brunton \& Kutz \cite{BruntonKutz19}, and some applications combining NNs linear mappings with non-linear modal interactions for turbulent flows are presented by Le Clainche {\em et al.}~\cite{LeClaincheetalJFM2022}.

\subsection{Regression and Classification}

Regression and classification belong to the  category of supervised learning. In supervised learning, labeled data sets are used, to guide  the machine during the training process.
The algorithm learns by minimizing the error with respect to the labelled target. Regression and classification are widely popular and very effective techniques, but 
can lead to erroneous results if data (or categories) are missing, which leads to problems during training and/or classification. 
The missing data or category can relate to a problem misconception leading to a category missing in the data, to data removal because of large deviations, to equipment malfunctioning, etc.  \cite{AizedAminSoofi_ArshadAwan_2017}. 
To avoid problems in these supervised-learning algorithms, a good a-priori knowledge (and analysis) of the problem at hand is advisable to ensure that the labelled training data is representative of the overall dataset.
In this section, we focus on the most common regression and classification algorithms, including: regression and logistic regression, decision trees, random forest, K-nearest neighbours, support-vector machines and 
Naive-Bayes classification.

A common task in supervised learning is regression, which is used to identify patterns and relationships within a dataset. In this task, the algorithm approximates a function for a continuous output ({\it e.g.,} a real value). Classic examples are linear (or quadratic) regressions, where linear (or quadratic) curves are fit to data. This type of regressions can be understood from a purely algebraic viewpoint, through least-square approximations ({\it i.e.,} Ridge regression using the L2 norm) but generalises to ML using NNs.

\begin{figure}[h!]
	\centering
	\includegraphics[width = \textwidth]{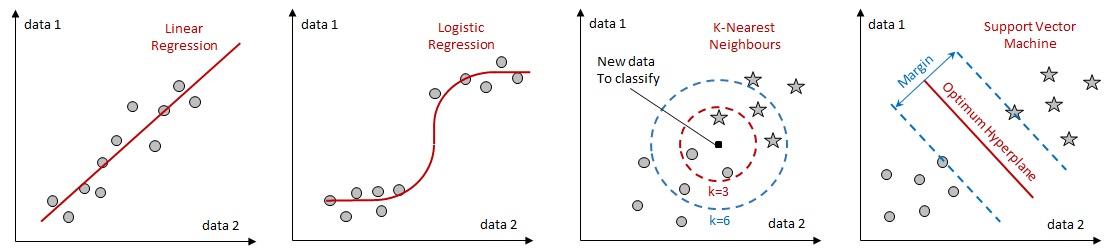}
	\caption{Summary of methods for regression and classification in machine learning. From left to right: linear regression, logistic regression, K-nearest neighbours and support vector machines.
 \label{fig_classification}}
\end{figure}

In linear regression with one variable, the  output $y$ will fulfill the equation of a straight line: $y=mx+c$, where $m$ is the slope, $c$ the intercept and $x$ the input.
The calculation of ($m$) and ($c$) will be based on the minimization of a cost function, {\it i.e.,} the minimizing the error between the training data and the  linear model. An example is shown in Fig.~\ref{fig_classification}. The regression model can be used to interpolate missing data or to extrapolate (forecast) outside the training-data range. 
When considering regression with multiple dependent variables, it is necessary to analyze the influence of each variable on the overall trend of the dependent variable (system output).
Multiple linear regression is a useful technique for more than one independent variable. This technique can achieves better fits than when using simple linear regression if multiple independent variables are involved.

Classification also belongs to the supervised-learning family. In this problem, the output will take a discrete set of values or categories. The algorithm will use pre-set labeled data to classify the data into multiple categories. According to the type of variables, we can distinguish categorical classification from numerical classification. 
Logistic regression is the simplest classification algorithm that performs a binary classification (between only two categories) and can help to determine the occurrence (or not) of an event.
The algorithm starts with data previously labeled in the two possible categories. These two categories are joined through a sigmoid curve, which will determine the probability that new data falls into one category or the other, see Fig.~\ref{fig_classification}. 
The sigmoid curve 
$$P(t)=\frac{1}{1+\exp{(-t)}}$$ helps classify the probability of success of an event; starting with a null probability (first category) and ending at a certain event (second category). The curve shows a change of curvature when the probability of success is 50\%. This point, called threshold, will differentiate the data classified into both categories, as depicted in Fig.~\ref{fig_classification}.


The decision tree is a conceptually simple algorithm, useful for classifying data.  
The algorithm uses a tree, in which each of the branches implies a decision leading to a classification. To make these decisions and determine which characteristics to discern, it is necessary to train it with data.
The construction of the trees, as well as the determination of the order of the different branches alludes to the reduction of dispersion (or entropy) in the data in each step providing an optimal tree with minimal dispersion.

Random forest \cite{Louppe} is an algorithm for both classification and regression tasks and is particularly useful for data sets in which there can be loss of information.
Its development starts from the construction of multiple decision trees (hence the concept of ``forest''). Each decision tree performs a classification task leading to the final result of the  random forest, which is the best classification of all trees.

K-Nearest Neighbours (KNN) is one of the simplest algorithms used for supervised classification. It classifies the input data based on the closeness to already classified data. The algorithm generates ``neighbors'' of data and  measures the distance (or ``similarity'') of new data with respect to the classified neighboring data. The KNN algorithm is effective if the data set is not excessively large (slow learning algorithm) and the data is free of noise.
The KNN concept can be understood by drawing a circle centered on the new data point, which  increases its radius until there are $K$ elements (belonging to a pre-set category) within the circle. $K$ represents the number of elements needed within the circumference to classify the model input; this gives the name to the algorithm K-nearest neighbors, as shown in Fig.~\ref{fig_classification}.
The choice of the parameter $K$ is essential to achieve the desired resolution of the problem, and needs to be large enough to avoid the influence of noisy data, but small enough to limit the computation time. A possible value is the square root of the labeled data ($n$), {\it i.e.,} $K=\sqrt{n}$.

Support-vector machines (SVMs) are similar to K-Nearest Neighbours and can be used to perform classification and regression tasks where the transformed feature space is very large. However, for large datasets, SVMs
suffer from high computational cost. 
SVMs create a hyperplane to delimit the neighbours or regions in the data set, see Fig.~\ref{fig_classification}. 
To determine the position of the hyperplane, the algorithm finds support vectors, defining the distance to the neighbours. This distance must be maximum for the classification to be successful.
Note that the hyperplanes may have various shapes: linear, quadratic, cubic, etc. 
SVMs can deal
with a wide variety of classification problems including high dimensions and non-linearities. A drawback is that SVMs require calibration of the parameters to attain good results.

Finally, the Naive--Bayes classifier has its mathematical basis in the conditional probability proposed by Bayes' theorem, whose formulation is: $$P(A|B)=\frac{P(B|A)\cdot P(A)}{P(B)}.$$
The logical operation is similar to the error back-propagation algorithm. The final probability will be based on the 
total probability taking into account the influence of the individual events on the following events (conditional probability). The total probability theorem, used to give the final output of the model is $P(B)=P(A_1)\cdot P(B|A_1)+P(A_2)\cdot P(B|A_2)+...$
The process requires to calculate the probability of each event individually and analyze its influence on the final response. This probabilistic analysis will finally lead to a decision.
The algorithm will predict events, based on present and past data, knowing the probability of occurrence of the predicted events.
Some advantages of Naive--Bayes classifiers are outlined in \cite{AizedAminSoofi_ArshadAwan_2017} and include: (i) they remain smooth to small changes in the network,
 (ii) they are flexible, allowing for transfer learning between classification and regression, (iii) they can handle missing data.

Gaussian-process (GP) regression is an extension of Bayesian-regression method that is used widely within the field of structural health monitoring.  Rather than predicting a single deterministic function, the predictive posterior is instead a distribution of functions that are consistent with the data that the model has been conditioned by. At any finite number of points in the output space, the distribution of these functions is Gaussian. Like SVMs, GP regression uses a \textit{kernel}, or \textit{covariance} function to estimate the similarity between two inputs and it is the selection of this kernel that defines the form that the posterior functions take, in combination with the optimization of the model hyperparameters. The GP can be defined by its mean and this covariance; often a zero mean prior is assumed, however, knowledge of the process can be incorporated into the model relatively simply using a \textit{mean function}. A commonly-used covariance function is the squared exponential: $$k(\bf{x},\bf{x}^\prime) = \sigma_f^2\;\exp\left(-\frac{1}{2\ell^2}\vert\vert \bf{x}-\bf{x}^\prime\vert\vert^2\right)$$, where $\bf{x}$ and $\bf{x}^\prime$ are the inputs to the model, $\sigma_f^2$ is the signal variance hyperparameter and $\ell$ is the lengthscale hyperparameter. The reader is referred to Ref. \cite{rasmussen} for an in-depth review of GP regression theory.

\subsection{Semi-supervised learning}

\begin{figure}[h!]
	\centering
	\includegraphics[width =  \textwidth]{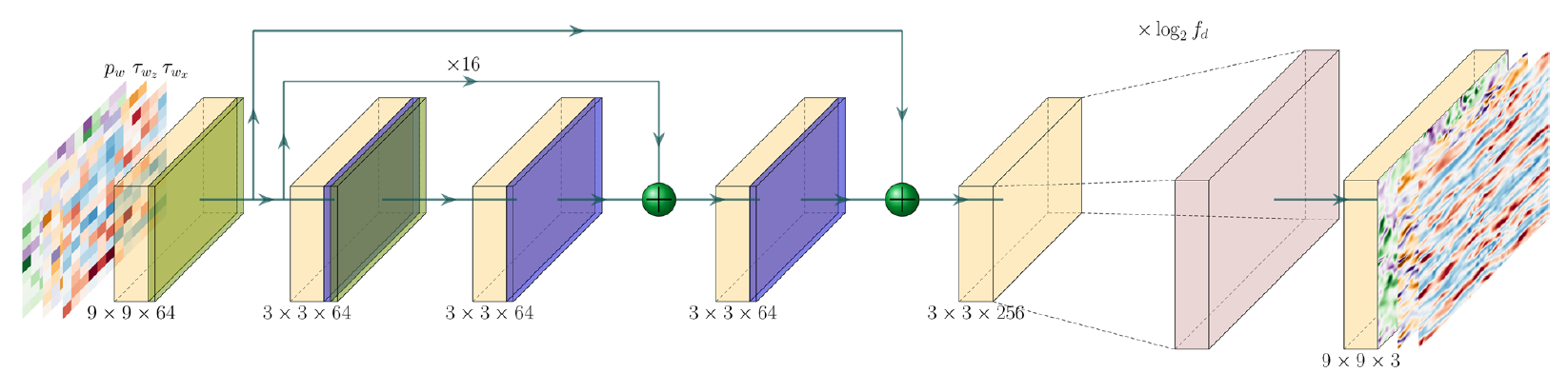}
	\caption{Schematic representation of a GANS architecture used to increase the resolution of the quantities measured at the wall in a turbulent channel flow. The color coding for each layer is 2D convolution (beige), parametric-ReLU activation (dark green), batch normalization (blue), sub-pix convolution (pink) and ReLU activation (light green). Note that the kernel size and the number of filters are shown at the bottom of each of the layers. Figure adapted from Ref.~\cite{guemes2} with permission of the publisher. 
 \label{fig:gans}}
\end{figure}

In this section we will discuss two types of semi-supervised algorithms: generative adversarial networks (GANs) and deep reinforcement learning (DRL). GANs~\cite{Goodfellow_et_al} are a type of generative model, which comprises two different parts: the generator ($G$) and the discriminator ($D$). These two parts of the network are assigned competing tasks, and they are trained against each other using game theory. A widely-used application of GANs is that of super-resolution, {\it i.e.,}, starting from a low-resolution input image $L_R$, the GAN architecture needs to produce a high-resolution version $\tilde{H}_R$ which should be consistent with the statistical features of the original high-resolution image $H_R$. In this setup, the generator will be tasked with creating the $\tilde{H}_R$ images given the $L_R$ input and the reference $H_R$. This process is illustrated for a turbulent flow in Fig.~\ref{fig:gans}. Note that this super-resolution task can also be conducted in an unsupervised manner by imposing physical properties on the data, as documented by Kim~{\it et al.}~\cite{kim_gans}. The discriminator is thus trained to differentiate $\tilde{H}_R$ and $H_R$, so that the generator gets progressively better at producing realistic images, and the discriminator improves its performance when identifying the ones that have been artificially generated by $G$. It is common to use a convolutional neural network (CNN) as generator, 
and the resolution increase is usually carried out at the end of the network by means of sub-pixel convolution layers~\cite{Ledig_et_al}. A common architecture for the discriminator in this type of task~\cite{guemes2} is to also use a CNN, followed by fully-connected layers, and using a sigmoid activation in the last one it is possible to obtain a probability value between 0 and 1 to discern whether the high-resolution image is real or not. 

\begin{figure}[h!]
	\centering
	\includegraphics[width = 0.5 \textwidth]{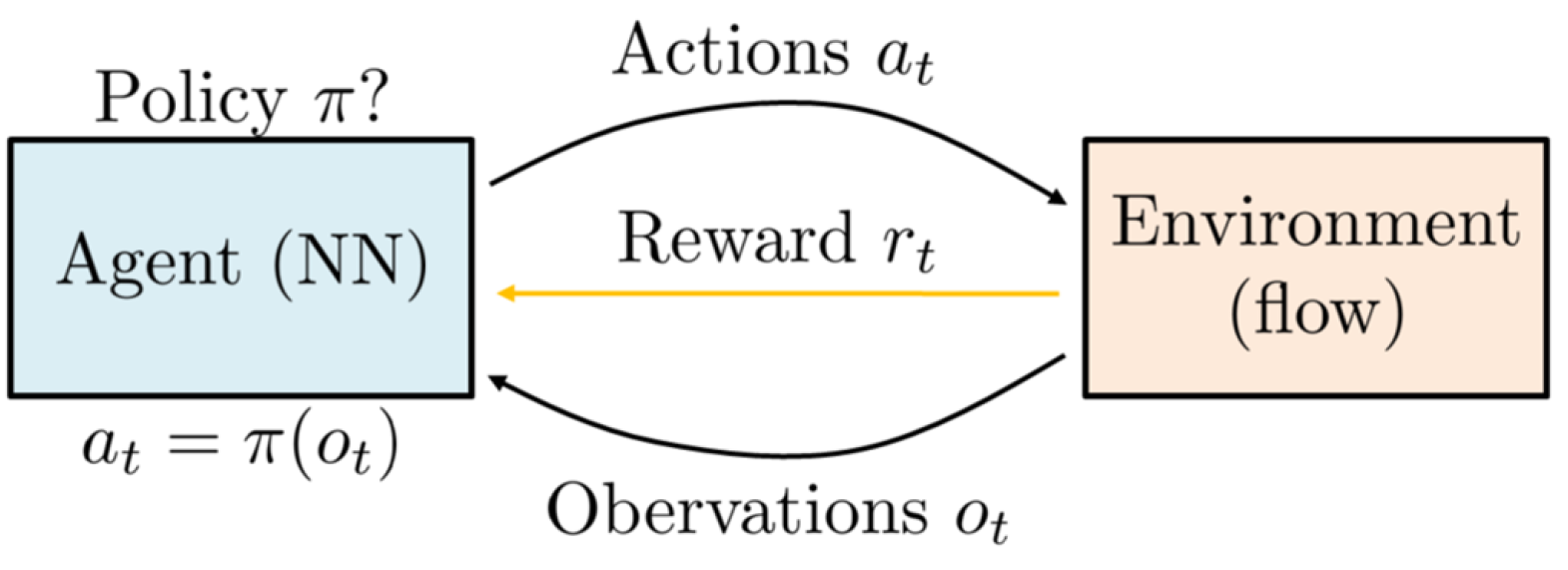}
	\caption{Schematic of the deep-reinforcement-learning process, where the aim is to find the optimal decision policy ensuring that the cumulative reward is maximized. Figure extracted from Ref.~\cite{drl_control} with permission of the publisher. 
 \label{fig:drl}}
\end{figure}

When it comes to DRL, in this framework an agent (which can be a neural network) performs actions on an environment (which would be the flow) in a closed loop. If we consider instant $t$, at which the environment has a certain state $s_t$, the agent receives a partial observation of that state denoted $o_t$. Based on this observation, the agent decides to apply a certain action $a_t$, which will have an impact on the system, modifying its state to $s_{t+1}$, and so forth. The quality of those actions under a certain norm is measured by the so-called reward $r_t$, which is provided periodically to the agent. Generally in reinforcement learning, the framework is aimed at developing an optimal decision policy $a_t=\pi(o_t)$ with the goal of obtaining the maximum cumulative reward over a certain time horizon. This process is summarized in Fig.~\ref{fig:drl}. One widely-used implementation is the proximal-policy-optimization (PPO) method~\cite{Schulman_et_al}, which has the advantage of being faster than {\it e.g.,} the trust-region policy optimization (TRPO)~\cite{Schulman_et_al}. There are other suitable options for DRL, such as deep-Q network (DQN) learning~\cite{Mnih_et_al}, which is very efficient mainly with non-continuous control. In this sense, it is important to thoroughly assess the dynamics of the problem at hand to choose the most suitable DRL implementation, see for instance the variations of DQN learning reported by Gu~{\it et~al.}~\cite{Gu_et_al}.



\subsection{Unsupervised learning: clustering and dimensionality reduction}

Compared to classification and regression approaches, unsupervised learning deals with unlabeled daya with the objective of "learning" the data structure without assistance.
Unsupervised methods allow processing large amounts of data, leading to the determination of the most relevant features that can be used for a lower-dimensional representation of the original data. As such, unsupervised learning techniques are highly relevant in aerospace applications.

\subsubsection{Clustering}
Clustering ({\it e.g.,} K-means, mean shift, Gaussian mixture models) \citep{bishop_2006}, is an unsupervised-learning method that does not require pre-sampled data to cluster regions and can automatically discover grouping within the data. An overview of clustering methods and comparisons for simple problems can be found in Ref.~\cite{10.5555/1953048.2078195}.  Here, we review the popular K-means, mean shift and Gaussian mixture models.

K-means is the simplest and faster clustering technique. The user inputs the number of clusters ``k'' and the algorithm initializes ``k'' centroids, assigning each point in the dataset to the closest centroid. Using these temporary clusters, new centroids are calculated employing the mean distance from all the data points. The process is repeated iteratively until the centroids do not move in the subsequent iterations. 

The mean-shift clustering method is a variant of K-means where instead of the centroid, a sliding circular window of radius “Kernel” are considered. The goal
is to slide the windows until the location that maximises the density of points inside the window is found. Density-based spatial clustering of applications with noise (DBSCAN) is a variant of mean-shift that considers noisy unclassified data, and can provide enhanced robustness. 

Gaussian-mixture models assume that the data follows a Gaussian distribution with an associated mean and standard deviation. The algorithm estimates the mean and variance of each normal distribution iteratively using the expectation-maximization method \citep{Dempster_1977} to provide the optimal estimation of these parameters.

There is some recent interesting in aerodynamics and CFD, which uses clustering to detect flow regions. For Reynolds Averaged Navier--Stokes (RANS) simulations, Saetta \& Tognaccini \citep{ettore_2022} and Lanzetta et al. \citep{lanzetta_2015} showed that various viscous sensors could be used as input to a clustering ML framework. The aim was to detect the boundary layer and wake regions for turbulent flow past aeronautical applications (airfoils, aircraft).  The results showed that the ML clustering outperformed the classic viscous sensors used to identify the flow regions in RANS. In the context of Large Eddy Simulations (LES), similar ideas have been proposed using Gaussian mixture to identify laminar and turbulent regions \citep{ferrer2022,Ferrer2022_2}, which can also be employed to perform local mesh adaptation to increase the accuracy of simulations.
Furthermore, Callaham et al. \citep{callaham_2021}  used unsupervised-learning techniques to identify the dominant physical processes for different flow scenarios and in Ref.~\citep{colvert_2017} the authors trained a neural network to classify different types of vortex wakes. 

\subsubsection{Dimensionality reduction}
One of the most common tasks in unsupervised learning is dimensionality reduction. Autoencoders use NNs (with linear or non-linear activation functions) to reduce the data dimensionality and  extract the main data features ~\cite{ae_modal}. In fluid dynamic problems, a very rich literature exists concerning the application of proper orthogonal decomposition (POD) \cite{LumleyPOD}. POD is also known with a variety of names, depending on the application field, {\it i.e.,} Karhunen-Loève decomposition (KLD) \cite{karhunen1947} in image and signal processing, Principal Component Analysis (PCA) \cite{Jolliffe1986} in the statistics literature, Empirical Orthogonal Functions \cite{lorenz1956}in oceanography and meteorology, etc. A detailed discussion on the relation between POD, PCA and KLD, and their connection with Singular Value Decomposition (SVD) is presented in \cite{WU20031103}. 

\subsubsection{Principal Component Analysis}

\begin{figure}[h!]
	\centering
	\includegraphics[height = 0.3\textwidth]{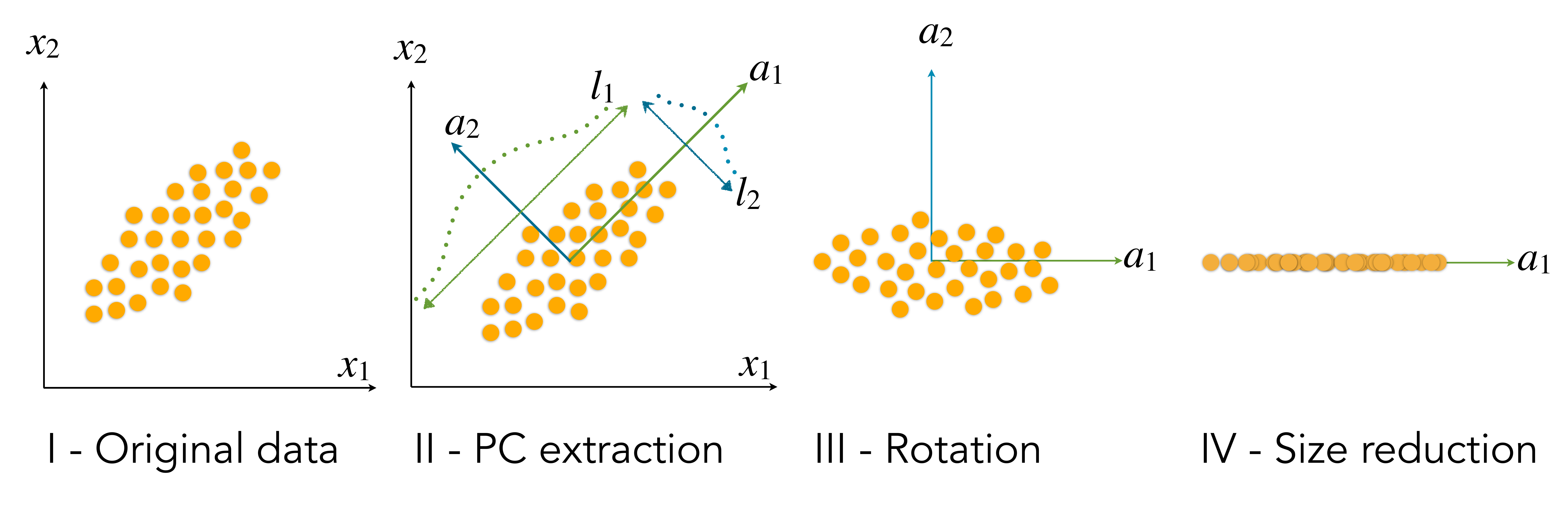}
	\caption{Steps in the dimensionality reduction process using PCA. \label{figPCA}}
\end{figure}

Dimensionality reduction techniques rely on the hypothesis that a reduced and optimal basis exists to represent the data. PCA thus finds a new, orthogonal basis to represent the data set, which is approximated by retaining only a subset of fewer directions, namely the ones accounting for most of the data variance in a transformed data set. Prior to performing PCA (as in all ML approaches) the raw data set is generally transformed (centered and scaled) to allow dealing with variables of different units and sizes, and to focus on fluctuations. Many options exist for data preprocessing and optimal choices cannot be defined universally but rather depending on the problem at hand \cite{Parente2013}. The eigenvalue decomposition of the covariance matrix yields the eigenvectors, denoted as the principal components (PCs), ordered by the magnitude of their corresponding eigenvalues. Truncating the PCs to a desired level of variance provides a low-dimensional representation of the original data (Fig. \ref{figPCA}), which is the best linear predictor of the data matrix $\mathbf{X}$ in terms of squared prediction error \cite{Jolliffe1986}.

PCs are determined to maximise variance, not interpretability. The complex structure of the PCs and the presence of non-negligible weights for most of the original variables on an individual PC might complicate the interpretation and exploitation of the PCs identified without supervision. To alleviate that, the eigenvector matrix can be rotated to aid physical interpretation of the scores, {\it i.e.,} the projection of the data onto the new basis. Many rotation methods exist, they can  either be orthogonal or oblique. Orthogonal transformations, such as Varimax rotation \cite{kaiser}, rotates the PCs rigidly while maximizing the sum of the variances of the squared coefficients within each eigenvector. The rotation generally leads to rotated PCs with few significant coefficients, thus aiding interpretation. After rotation, the amount of variance accounted for by a set of PCs is conserved but the proportion is redistributed amongst the components. 

\subsubsection{Local dimensionality reduction}
\begin{figure}[h!]
	\centering
	\includegraphics[height = 0.3\textwidth]{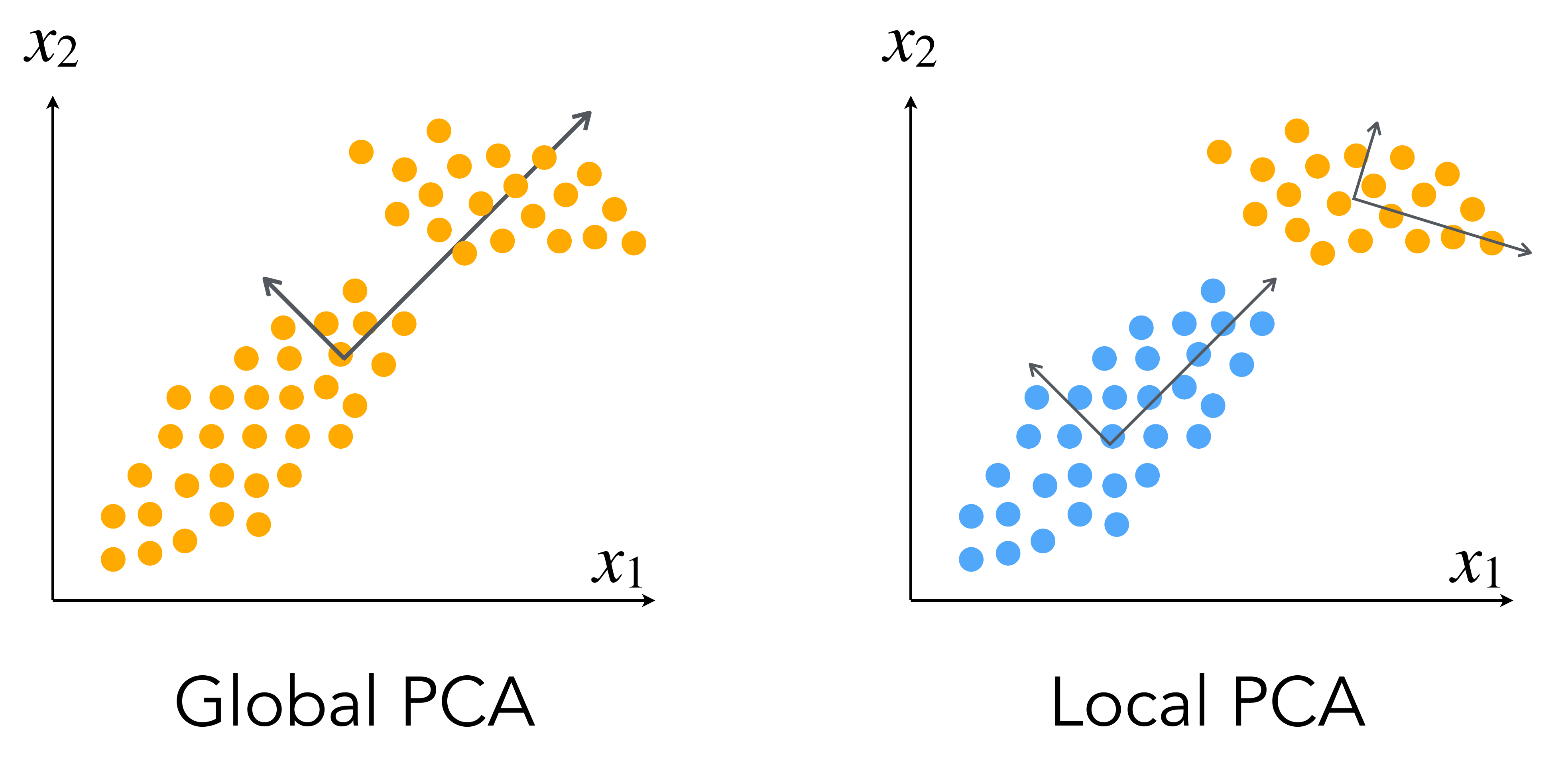}
	\caption{Schematic representation of global and local PCA. \label{figLPCA}}
\end{figure}

Typical flow data sets relevant to aerospace applications display non-linear relationships among variables. PCA is a multi-linear technique, {\it i.e.}, additional PCs are needed to approximate the original data set with low reconstruction error to compensate for the data non-linearity. To overcome the limitations of global PCA, an alternative is to identify local principal curves \cite{localcurves} and locally linear embeddings (LLE) \cite{Roweis2000}, and to perform local modal decompositions, i.e., applying PCA locally (LPCA).

local PCA (LPCA) can be exploited, partitioning the data into clusters by either using an {\it a priori} chosen feature, or using a Vector Quantization (VQ) technique coupled to the local application of PCA within clusters. The resulting approach, denoted as VQPCA \cite{KambhatlaL97, Parente2011}, assumes that the non-linear manifold can be locally approximated by a linear one as the data manifold will not curve too much over the extent of the local region. In the first case a supervised clustering is carried out, while the latter is fully unsupervised. In VQPCA, the data are assigned to the different clusters by minimising the reconstruction error of each point to a given set of clusters (Fig. \ref{figLPCA}). The algorithm is iterative and convergence is achieved when the centroids' change from one iteration to the next one is below a prescribed threshold.

\subsubsection{Kernel PCA}
Kernel PCA (KPCA) is a non-linear dimensionality reduction technique that makes use of kernel methods. In KPCA \cite{kernelpca}, the original data-set is transformed into an arbitrarily high-dimensional feature space where the relationship amongst the variables is considered to be linear and PCA can be effectively carried out. A non-trivial, arbitrary function is chosen but never calculated explicitly, which allows using high-dimensional feature spaces. The non-linearity of the KPCA method comes from the use of a non-linear kernel function that populates the covariance matrix (kernel trick). Thus, the eigenvectors and eigenvalues of the covariance matrix in the feature space are never actually solved. The choice of the kernel function modeling the covariance between two different points in the feature space is up to the designer. An example is the choice of a Gaussian kernel. Since no operations are
directly carried out in the feature space, KPCA does not compute the PCs themselves, but the projections of our data onto those components, namely the scores.

One of the main drawbacks of kernel PCA is that PCA is performed on a covariance matrix that scales with the number of observations rather than the number of variables, thus increasing the computational burden of the method. In addition, no straightforward operation is available to reconstruct the data from the KPCA scores, since the kernel is never computed explicitly. This is in contrast to linear PCA, where a direct mapping exists. To reconstruct the data in Kernel PCA, a minimisation problem must be solved, making the process computationally more expensive compared to PCA.

\subsubsection{Autoencoders}
An autoencoder (AE) is an unsupervised neural network (NN) aiming at learning the reduced representation (encoding) of a dataset. Along with the encoder, that projects the data into a non-linear manifold (codes), a decoder reconstructs the data from the codes side, minimising the reconstruction error. The simplest form of an AE is a feed-forward, non-recurrent neural network having an input layer, an output layer and one or more hidden layers connecting them, where the output layer has the same number of neurons as the input layer, and with the
purpose of reconstructing its inputs (minimizing the difference between the input and the output) via a loss function that can  accommodate regularization and sparsity terms. It is worth noting that a shallow AE with linear activation functions corresponds to PCA.





\section{Fluid mechanics\label{sec:FluidDyn}}


Fluid mechanics is closely connected with the aerodynamic performance of airplanes, and it is therefore essential to properly model and predict the properties of the flow around the aircraft in order to carry out optimization and design tasks. Very close to the airplane surface a thin layer, called boundary layer~\cite{clauser}, develops. This region plays a critical role in aircraft design, and it is extremely challenging to model, particularly at high Reynolds numbers, {\it i.e.,} when the boundary layer is {\it turbulent}~\cite{marusic_review}. 

\subsection{Computational fluid dynamics}

The numerous success stories in ML documented for a wide range of areas~\cite{vinuesa_nat} over the past years motivate the possibility of using ML methods for simulating fluid flows. An overview of how ML can help computational fluid dynamics (CFD) is provided by Vinuesa~and~Brunton~\cite{vinuesa_brunton}, who essentially identified potential in many areas including accelerating CFD and improving turbulence modeling. It is important to note that ML methods will not constitute a replacement for traditional CFD methods~\cite{godunov1959finite,canuto}, which are based on discretizing the governing equations and integrating them numerically. Instead, it is important to identify concrete areas within CFD that can benefit from ML. For instance, interesting work has been developed by Kochkov~{\it et al.}~\cite{hoyer} on improving the accuracy of coarse simulations by means of deep learning in simplified flows. Other areas of potential are the efficient resolution of the Poisson problem~\cite{poisson3}, which is encountered in operator-splitting methods, or approaches to reduce the size of the computational domain through efficient inflow~\cite{fukami_inflow,yousif_inflow} and far-field~\cite{morita_et_al} conditions. Manrique de Lara and Ferrer~\cite{deLaraFerrer1,deLaraFerrer2} have used NNs to accelerate high-order simulations by including a corrective forcing that achieves high accuracy when running a low-order case, providing a framework for accelerated simulations. When it comes to turbulence modeling, Duraisamy {\it et al.}~\cite{duraisamy_et_al} provided a review of the potential of ML to improve RANS models. Note that these models, which are based on Reynolds averaging and typically rely on an eddy-viscosity approach~\cite{boussinesq} to represent the Reynolds stresses, are not necessarily satisfactory when simply using high-fidelity data to address the closure problem~\cite{wu_et_al_2018}. The anisotropy tensor, which is an important component of RANS models, was predicted by Ling {\it et al.}~\cite{lingetal} using a deep-learning architecture which embedded Galilean invariance, an approach leading to excellent results. On the other hand, Wu {\it et al.}~\cite{wu_et_al_rans} proposed combining the physical knowledge from the flow with deep-learning architectures, producing a framework that allowed predicting both linear and nonlinear parts of the Reynolds-stress tensor separately. In addition to improved interpretability, this approach led to very good results in a number of configurations. The interpretability~\cite{rudin} of the deep-learning models for RANS simulations was one of the main features of the framework proposed by Jiang {\it et al.}~\cite{jiang_et_al}. When it comes to interpretability, we would like to highlight that deep-learning models are generally able to provide accurate predictions based on the input data, but it is typically not possible to establish the concrete relationships relating input and output. An interpretable model allows to establish such as connection, and there are a number of research directions~\cite{vinuesa_sirmacek} starting to enable this property in deep-learning models. A good example of this, which relies on genetic programming, is the work by Cranmer {\it et al.}~\cite{cranmer_et_al}. It is also worth noting that Weatheritt~and~Sandberg~\cite{sandberg1} developed a methodology for RANS modeling based on gene-expression programming~\cite{gep} which essentially produces models that are interpretable since they rely on combinations of basic symbolic expressions. Their models exhibit excellent results in a number of complex geometries where RANS models typically fail, including cases with secondary flows and significant three-dimensionality. Along these lines, a generelization of the constitutive relations used to model the Reynolds stresses can also help to properly predict complex flows~\cite{spalart}, and in this sense data-driven model discovery is a promising research direction.

Another simulation area where ML has shown potential is that of large-eddy simulation (LES). As opposed to RANS, where all the turbulent scales are  modeled  through the Reynolds stresses, in LES a certain range of (the larger) scales is simulated, and low-pass filtering is employed. The scales with higher wavenumbers ({\it i.e.,}, with smaller wavelenghts) are then modeled through a so-called subgrid-scale model (SGS). There are several deep-learning-based methods to develop such SGS models, {\it e.g.,} the approach based on CNNs by Beck {\it et al.}~\cite{beck_et_al}. CNNs are wiely used in computer vision, and they can exploit the spatial correlations present in the data to enhance their predictions. Thus, they are suitable for making accurate predictions in the context of turbulent flows. Another CNN-based approach was proposed by Lapeyre {\it et al.}~\cite{lapeyre_et_al}, in their case applied to reactive flows. Despite the potential of these approaches, in principle they rely on data of higher fidelity ({\it e.g.,} direct numerical simulations, DNSs) to train the deep-learning models. The higher-fidelity data is used to develop an SGS model able to reproduce the behavior of the non-modeled simulation in the LES case. The limitation of this approach relies on the need of such high-fidelity data, which limits the applicability of these models in a wide range of cases beyond those considered in the training stage. An alternative to this approach was proposed by Novati~{\it et al.}~\cite{petros_natmi}, who used reinforcement learning (RL) 
to develop an SGS model. The advantage of this method compared with the one above is that it is unsupervised, {\it i.e.,} it does not require the high-fidelity data as a reference. Furthermore, it is able to develop SGS models that do not necessary emulate the behavior of a high-fidelity simulation, but rather establish a condition consistent with the coarse case being simulated. This in principle can favor the generalizability of such LESs. Another area of potential interest within ML is that of wall-model development. Wall models are employed at very-high-Reynolds-number applications, where the whole near-wall region is replaced by a suitable boundary condition~\cite{moeng}. Note that at the moment there is not a clear approach to accomplish this for general cases, given the fact that the flow in the outer region (for example streamwise pressure gradients) significantly determines many of the flow features closer to the wall~\cite{wing1M,pozuelo}. Another difficulty is the fact that in wall-bounded turbulence over roughness the outer region has no way of knowing whether the wall is smooth or rough, and therefore there is uncertainty regarding the best approach to set these boundary conditions~\cite{jimenez_review}. Despite these limitations, important progress has been made based on data-driven methods relying on physical arguments~\cite{mizuno,kenzo}. Deep learning is also providing streategies to be able to map the flow in the outer region and at the wall-normal location where the boundary condition would be set, either through direct prediction~\cite{ari_etmm} or by determining a virtual velocity~\cite{moriya}. Another promising approach relying on ML is that developed by Bae~and~Koumoutsakos~\cite{bae_koumoutsakos}, which is based on reinforcement learning to determine the correct boundary information close to the wall.

\subsection{Reduced-order models}

Fluid-mechanics systems involving turbulence are high dimensional and chaotic, thus it is convenient to develop frameworks to reduce their dimensionality and simplify their analysis. The so-called ROMs are not only relevant when it comes to shedding light on the physics of these systems, but also regarding other applications such as flow control or optimization~\cite{Taira2017aiaa}. Classical methods for ROM development rely on linear algebra, where a widely-used approach is the so-called proper-orthogonal decomposition (POD)~\cite{lumley}, which relies on the singular-value decomposition (SVD) as discussed above. The POD framework, which is essentially data driven, is applied to $N+1$ three-dimensional snapshots of the flow field $\bm{u}(\bm{x},t)$, where boldface denotes tensors, $\bm{u}$ is a vector containing the three velocity components at instant $t$ and $\bm{x}$ is a vector containing the spatial coordinates. The POD of the spatio-temporal signal $\bm{u}(\bm{x},t)$ would yield:
\begin{equation}
\bm{u}(\bm{x},t)=\bm{\overline{u}}(\bm{x})+\sum_{i=1}^{N} a_i(t) \bm{\phi}_i(\bm{x}), 
\label{eq:pod}
\end{equation}
where $\bm{\overline{u}}(\bm{x})$ is the mean flow, $a_i(t)$ are the temporal coefficients and $\bm{\phi}_i(\bm{x})$ are the spatial modes. Consequently, POD separates spatial and temporal information in the flow, and because the spatial modes are ranked by their energy contribution to the original signal~\cite{lumley}, it is possible to truncate the sum in eq.~(\ref{eq:pod}) at a given number of modes, defining a certain level of energy reconstruction. Another interesting related technique is the so-called dynamic-mode decomposition (DMD)~\cite{Schmid2010jfm}, which also identifies a low-dimensional model of the original system, with modes ranked by frequency (instead of energy). These methods are linear, and since turbulent flows are inherently non linear, it may be interesting to to explore methods which include and exploit such non linearity. An example of this is the so-called higher-order dynamic-mode decomposition (HODMD)~\cite{LeClaincheVega17}, which embeds non-linearity in the temporal dependencies among the snapshots used to develop the ROM. In this sense, ML has great potential to also help develop suitable ROMs able to exploit the massive amounts of data available and the non-linearities characterizing these flows. An early example of this is the approach by Milano~and~Koumoutsakos~\cite{milano_koumoutsakos}, who employed deep neural networks to learn the temporal dynamics of ROMs of the near-wall region of turbulent channels. In fact, they established meaningful connections between neural networks and POD, namely the fact that when nonlinearities are removed from neural networks one recovers POD when developing ROMs. Other types of ROMs purely based on HODMD (a tool generally used in fluid dynamics to identify the main flow physical patterns) have been used to predict the temporal evolution of numerical databases in transitional~\cite{LeClainche19,LeClaincheVegaPoF17} and turbulent flows~\cite{LeClaincheFerrerEnergies18}. The advantage of these models is that they were based on physical principles (driving the main flow dynamics), although the type of models used were purely linear, limiting their application in some case. Recently, Le Clainche {\em et al.}~\cite{LeClaincheetalJFM2022} combined HODMD with  NNs  to develop a model capable to predict the wall shear stress in a channel with a porous wall leading to drag reducing/increasing effects (anisotropic turbulence). The model used HODMD to identify the main instabilities leading the flow dynamics. The non-linear interaction of the HODMD (also known as DMD) modes was also considered in the model to then combining this information (related to the flow physics) with simple architectures of NNs to predict the wall shear stress for long time intervals. Based on the same idea of developing predictive ROMs containing information modelling the flow physics, Abad{\'i}a-Heredia {\em et al.}~\cite{AbadiaHerediaetalESWA2022} combined POD modes with NNs using CNNs or LSTM architectures to predict the temporal evolution of the flow using data from the transitory of CFD simulations. In all these  examples, using data dimensionality techniques based on the physical principles describing the flow dynamics,  allowed to use one-dimensional architectures for the NNs. L{\'o}pez-Mart{\'i}n {\em et al.}~\cite{LopezMartinetal2022} also show that using both CNNs or LSTM it is also possible to accurately predict the temporal flow evolution from transient CFD databases, but using three-dimensional NNs architectures instead.

The low-order model of the near-wall cycle of turbulence proposed by Moehlis {\it et al.}~\cite{Moehlis}, which exhibits all the relevant phenomena present in the high-dimensional data, has been used as a platform to assess the feasibility of ML techniques to predict the temporal dynamics of the flow. For instance, LSTM networks were used by Srinivasan {\it et al.}~\cite{srinivasan2019predictions} to model the system, obtaining excellent predictions of turbulence statistics, Poincar\'e maps and Lyapunov exponents (following system perturbation). Other successful approaches to model the temporal dynamics in Ref.~\cite{Moehlis} include the Koopman-based framework (with non-linear forcing) by Eivazi {\it et al.}~\cite{eivazi_9} (which can provide in some cases better short-term predictions than the LSTM at a reduced cost) and the technique based on physics-constrained reservoir computing by Doan {\it et al.}~\cite{doan_et_al} (which is able to accurately model extreme events). 
Borrelli~{\it et al.}~\cite{borrelli_et_al} employed deep learning to predict the temporal dynamics of a minimal turbulent channel, achieving very good results, in line with those achieved for low-order models of turbulent flows. Finally, AEs are another very promising method for modal decomposition, because they enable developing non-linear ROMs. CNNs have been employed for AE-based modal decomposition~\cite{murata2020nonlinear,fukami_autoencoders}, and Eivazi~{\it et al.}~\cite{ae_modal} have shown their applicability to complex turbulent flows. In fact, they showed significant reductions of the number of modes necessary to reconstruct the original system compared with POD, while retaining orthogonality of the modes. 

\subsection{Experiments}

Experiments can also benefit from the numerous developments in the ML field. In general, exploiting the vast amount of data available in numerical simulations to improve experimental procedures is a very promising endeavor. Kim~and~Changhoon~\cite{kim_heat} have shown that the measured wall-shear stress can be used to predict the heat flux in a wall-bounded flow by means of deep neural networks, thus paving the way to augmented experimental data when no thermal quantities are measured. Another important area within experiments is that of non-intrusive sensing, {\it i.e.,} the possibility to measure at the wall and predict the flow behavior farther from the wall. This also has important applications in the context of flow control, as discussed below. Guastoni~{\it et al.}~\cite{guastoni_et_al} showed that fully-convolutional neural networks (FCNs) 
can be used to predict the velocity fluctuations above the wall in a turbulent channel, using as input the wall-shear-stress components and the wall pressure. Note that in FCNs all the layers throughout the network are based on convolutions, whereas this restriction does not apply in the more general CNNs. This is a very important result, because the wall quantities are easier to measure without disrupting the flow. Note that they obtained excellent predictions in the near-wall region, although their quality deteriorates farther away. Furthermore, they showed that deep neural networks are more suitable than traditional prediction methods such as linear stochastic estimation (LSE)~\cite{miguel} or extended POD (EPOD)~\cite{boree} to predict turbulent flows. G\"uemes~{\it et al.}~\cite{guemes} proposed an extension to the FCN approach by combining it with POD, such that the neural network is used to predict the temporal coefficients of the POD modes of the flow on a certain plane. Later, G\"uemes~{\it et al.}~\cite{guemes2} extended this work to the implementation of generative adversarial networks (GANs), which are able to perform super-resolution tasks. The GANs-based approach enabled performing accurate and robust predictions from few sparse measurements at the wall, as opposed to the previous studies, which required finely-resolved wall inputs. Another important framework for experiments is that of physics-informed neural networks (PINNs)~\cite{raissi2019physics}. This is a framework which exploits the machinery around the training process of neural networks (mainly the automatic differentiation) to formulate and solve partial differential equations (PDEs). Note that PINNs are currently being used to augment experimental measurements~\cite{raissi2020science} and to improve the quality of the measurements in fluid-mechanics problems~\cite{pinns_exp,pinns_etmm}. It is important to note that experiments in fluid mechanics rely on a number of empirical corrections, where data-driven methods can help to develop more general correction schemes~\cite{enhancing,role}. And as mentioned above, ML is also helping to develop more robust flow-control strategies. Non-intrusive sensing, particularly combined with opposition control~\cite{choi94}, is an excellent tool to enhance the experimental implementation of flow-control algorithms. Other data-driven methods for flow control include Bayesian regression based on Gaussian processes~\cite{rasmussen}, as done in the context of boundary-layer control by Mahfoze~{\it et al.}~\cite{Mahfoze}, and genetic programming, which has been recently used to control several complex flows~\cite{Li_et_al,Minelli_et_al}. Perhaps the most promising approach for control of fluid flows is DRL, which is a data-driven technique capable to discovering novel and more effective control solutions, as proved by Rabault~{\it et al.}~\cite{rabault2019artificial} in the case of the two-dimensional flow around a cylinder. Application of DRL to redue the length of separation bubbles has been documented by Guastoni~{\it et al.}~\cite{drl_bubble}, and an overview of DRL applications to turbulent flows, including flow control, has been provided by Vinuesa~{\it et al.}~\cite{drl_control}. 
 
\section{Aerodynamics\label{sec:Aerodyn}}

Improving aerodynamic design in aircraft is one of the main goals of the aerospace industry.  Aircraft wings, airfoil and turbine engine blades are some of the aircraft parts that need to be studied in detail to provide more efficient designs reducing drag (minimizing pollutant emissions) and noise (diminishing airframe and engine noise). The aerospace industry is already concerned about some design concepts that could result in more efficient designs reducing noise and drag. For instance, high aspect ratio wings with adaptive sections are connected to reduced vortex and wave drag in the aircraft, while to reduce friction and pressure drag it is necessary to develop flow control strategies to promote laminar flow, reducing the drag of the turbulent skin friction and controlling the unsteady flow separation and turbulent noise source. To reduce critical loads in the aircraft, in addition to control flow separation, it is important to control the presence of shock waves and unsteady loads for instance delaying the onset of buffeting, preventing flutter~\cite{MendezetalAST2020}, which is induced by linear aerodynamics, or controlling other aeroelastic responses leading to limit cycle oscillations (LCO), which are induced by aerodynamic non-linearities~\cite{ValeroAST2013}. Optimization, flow control and aeroelasticity are challenging topics that the current aerospace engineering industry need to enface during the next upcoming years to be able to develop more efficient, less pollutant and more silent aircraft designs. 

CFD numerical simulations and wind tunnel experiments are generally carried out to extract information connected to the aircraft performance, with the aim at using such knowledge to improve aerodynamic design. Aerodynamic data gathered from numerical and experimental sources can be divided in two groups: (i) distributed data, which includes skin friction and surface pressure distribution, and (ii) integral quantities, which include drag, lift and moment coefficients~\cite{KouArxiv2022}. However, the high cost associated to the performance of numerical simulations (large computational times and memory resources) and experiments (elevated monetary cost), encourage researchers to use aerodynamic models to solve design and multiphysics problems.

The lack of accuracy in classic aerodynamic models, which rely on empirical laws and classical theory (potential flow), combined with the rapid advance in data science and the current possibilities to generate and store large amount of high-fidelity numerical and experimental databases, motivate researchers to develop novel strategies to develop data-driven aerodynamic models. These models are more reliable and accurate than theoretical models and capable to provide relevant information connected to the physical insight of the problem under study, which opens new possibilities for applying the data-driven models to new flow control, optimization, aeroelasticity, flight dynamics and flow reconstruction applications~\cite{KouPAS2021}.  

Aerodynamic data-driven models can be divided into three groups: (i) semi-empirical models, which obtain the drag, lift and moment coefficients of airfoils as the result of linear or non-linear differential equations in charge of describing the flow dynamics, (ii) surrogate models, which are generally used in steady applications to find unknown variables using as general input geometrical parameters or flow conditions ({\it i.e.,} Mach number, Reynolds number, …), and (iii) ROMs, which include models for feature extraction ({\it i.e.,} POD~\cite{sirovich}, DMD~\cite{Schmid10}) that are generally used in large dimensionality flow datasets, and models for system identification based on input-output aerodynamic data ({\it i.e.,} neural networks, Kriging~\cite{GlazFriedman2010,Ronchetal2011}, Volterra series ~\cite{Raveh2001,Marzocca2004}, other linear and non-linear models such as Eigensystem Realization Algorithm -ERA , Auto-Regressive with Exogenous input -ARX, Non-linear ARX, Auto Regressive Moving Average -ARMA, Non-linear ARMA with exogeneous input -NARMAX, et cetera).

During the last years, AI, ML and more specifically, NNs, have emerged as highly potential tools to improve and transform engineering design and analysis, offering flexible, robust and versatile tools to develop ROMs and surrogate models, both suitable for finding optimal solutions in aerodynamic optimization~\cite{SunWang2019,LI2022100849}. ML applications to improve aerodynamic design solving optimization problems are multiple. For instance, NNs have been used to develop models (i) capable to accelerate CFD simulations, which increases the optimization efficiency since it is possible to calculate the integral quantities and distributed data of several design geometries in a short time simplifying the possibilities of finding the optimal designs~\cite{Lulianoetal2016,Nguyenetal2015}, (ii) also these models can be used to reconstruct high-fidelity data from sensor measurement ~\cite{Kanietal2017} or (iii) to provide a general idea about the proper combination of aircraft weight, wing area and engine thrust helping in the early stages of aircraft design to take decisions~\cite{Oroumiehetal2013}. ROMs can be combined with other analytical tools, can predict the aerodynamic loads and can represent aerodynamic systems, allowing several types of analysis at a reduced computational cost~\cite{KouZhang2016}. 

Even though ML is an emerging science in most of the subject connected to aerospace engineering, NNs have been used since the 90s to model aerodynamic loads for flight dynamics ~\cite{LinseStengel90}. In the recent extensive and detailed review of Kou \& Zhang~\cite{KouPAS2021}, the authors present several data-driven modelling methods applied to improve aerodynamic performance, including several references related to NNs applications since the 90s. The authors conclude as a general observation for the field of data-driven aerodynamic modelling that the future trends in this field should focus on developing novel data fusion techniques, allowing to combine knowledge from numerical simulations with different fidelities and experiments, maximizing in this way the capabilities of the models developed. The authors also reveal the need of developing new models less dependent on data and adding information about the flow physics, which includes interpretability on the results. The need of solving more complex applications, such as turbulent flows, complex geometries, industrial problems, three-dimensional and multi-scale flows ({\it i.e.,} massive flow separation), etc., encourage researchers to find novel alternatives to develop non-linear accurate models to improve aircraft designs. Evaluating the real applicability of these models by means of using uncertainty quantification tools, is critical for the industry of the future. These conclusions, generalized for data-driven modelling, can also be particularized to extend, and improve ML and NNs applications to solve aerodynamic problems. 

\subsection{Aerodynamic coefficients estimation}

The performance of the ML model depends on the type of identification method, the model structure and the type of data analysed. Flight testing, experiments and numerical simulations are different data sources to obtain unsteady aerodynamic forces. Using ML tools and considering data fusion techniques, which allows to combine data from different fidelities and types (experimental and numerical), would contribute to maintain the proper balance between cost of data generation and model accuracy ~\cite{KouPAS2021}. However, most of the models, only consider data coming from a single source. In this line, it is possible to mention the representative work of the following authors, ordered in chronological order since the 90s, who applied NNs for the identification of aerodynamic forces (lift and drag). More specifically, Linse \& Stengel~\cite{LinseStengel90} used NNs to estimate the aerodynamic coefficients, presenting a novel methodology that was based on extracting additional training information about the first partial derivatives of the aerodynamic coefficients. The method was successfully tested in flight test data generated numerically, solving a non-linear problem of twin-jet transport, and used for non-linear control of aircraft. Suresh {\em et al.}~\cite{Sureshetal2003} used RNNs to identify the dynamic stall effect of rotor blade on an OA212 airfoil. The authors predicted the lift coefficient in a case at high angle of attack using experimental databases and intended to show the generalization capabilities of the method presenting a way to generalize the error to model training. More recently, Secco \& Mattos~\cite{SeccoMattos2017} developed a surrogate model using NNs to predict aerodynamic coefficients of transport airplanes with high accuracy. The NN-based model used data from about $100.000$ cases calculated with CFD numerical simulations and the flight condition, airfoil geometry and wing planform were used as an input. The computational time to estimate the aerodynamic coefficients was $\sim4000$ times smaller than the time required by the CFD solver. This model was also successfully used for optimization design. Recently, Hou {\em et al.}~\cite{ Houetal2019} used NNs to study the effect of aerodynamic disturbances (due to manoeuvres or gusts) on the wing surface using pressure measurements. The database was generated by an inviscid vortex method in a two-dimensional flat plate. NNs are used to estimate the angle of attack and the leading-edge suction parameter. The authors first combine CNNs and RNNs providing an accurate estimation these variables. In a second approach, a novel and more accurate approach is presented, where ML is integrated in a dynamical system framework to further correct the prediction using a smaller number of parameters. The authors also show that adding random noise to the database it is possible to prevent overfitting. In the same year, to overcome the limitations in adapting input variables with different orders of ROMs, Wang {\em et al.}~\cite{Wangetal2019} proposed using a fluzzy scalar radial basis function NN, which combines fuzzy rules and radial basis function neural networks (RBFNNs), enhancing in this way, the generalization capabilities of the model to complex dynamical systems. This model was successfully tested to estimate the unsteady aerodynamics in transonic flow airfoils. RBFNNs were previously used by Ghoteyshi {\em et al.}~\cite{Ghoteyshietal2013} to compute the unsteady aerodynamic loads in an airfoil with pitching and plunging motion.

\subsection{Aeroelasticity}

Studying in detail aerodynamic linear and non-linear responses, for instance producing flutter (linear) or leading to LCO (non-linear), is extremely important to reduce critical loads in the aircraft. Non-linear dynamical modelling is a challenging task, since instabilities can be present in both training and simulation, additionally, purely non-linear models sometimes are unable to identify the main dynamics in systems with dominant linear behaviour. The possibilities of using NNs to combine both, linear and non-linear effects to develop models, have motivated several authors to focus their work on developing ML-based models to estimate aeroelastic responses in the aircraft. To name a few, Marques \& Anderson~\cite{MarquesAnderson2001} used a finite memory NN to model the nonlinear transonic aerodynamic aircraft response, motivated by the modelling difficulties generally found to control aeroelastic responses. The NN model used a supervised training process with multiple input-output sets from numerical data solving Euler equations. The finite memory NN architecture allowed to identify the memory in the hidden layer of the finite impulse response. Zhang {\em et al.}~\cite{Zhangetal2012} developed a non-linear unsteady aerodynamic model using RBFNNs employing the aeroelastic system self-excited vibration signal as input. The non-linear aerodynamic fluid-structure interaction (FSI) model was coupled with the structural equations of motion, accurately estimating the LCO change with velocity in a transonic flow with large shock motions. Mazhar {\em et al.}~\cite{Mazharetal2013} used NNs to successfully interpolate the pressure loads on unmanned aerial vehicles (UAVs). The aim was to present an accurate model suitable for UAV structural design, which was based on one-way FSI. The performance of NNs was superior compared to standard interpolation techniques using high order polynomials. Mannarino \& Mantegazza~\cite{ MannarinoMantegazza14} used recursive NNs to develop a model using high-fidelity numerical data solving a non-linear aerodynamic model. The NN model was successfully used to determine aeroelastic limit cycles using two different approaches: (i) through time marching simulations and (ii) through direct time collocation. The model was successfully tested in the Benchmark Active Control Technology (BACT) wing~\cite{ Riversetal92}, which consist o a rectangular wing pitching around its mid chord axis and with a NACA0012 airfoil section. To model non-linear unsteady aerodynamics, both at constant and varying flow conditions, Kou \& Zhang~\cite{KouZhang2017} successfully tested the performance of multi-kernel NNs, which are able to improve the generalization capabilities of the model. They also succeeded in using recursive NNs to predict non-linear aeroelastic behaviours and aerodynamic forces in a NACA 64A010 airfoil for multiple flow conditions in transonic flow. More recently, Li {\em et al.}~\cite{Lietal2019} used NNs, more specifically, LSTM networks to develop a model capable to predict non-linear unsteady aerodynamics, which was successfully tested for various Mach numbers on a NACA 64A010 airfoil that was pitching and plunging in a transonic flow. The model accurately captured the aerodynamic and aeroelastic capabilities of the dynamical system, showing the generalization capabilities of this NN architecture. LSTM architectures are perfect to capture the time-delayed effects of unsteady aerodynamics without the need of including any additional information about the orders selection (in unsteady aerodynamics it is necessary to consider the time-delayed effects from the input-output dynamical system). 

\subsection{Design optimization}

Design optimization is also a highly important discipline in aerospace engineering, where machine learning has recently gain strong relevance in the field. Aerodynamic design is related to airplane wing, airfoil geometry, UAVs, turbine engine blades, flight conditions, etc.. Optimization should be smart, efficient, and capable to easily balance design space ({\it i.e.,} shape, morphed geometry, etc.) and modelling efforts. In the recent review article by Sun \& Wang~\cite{SunWang2019} the authors summarize the main contribution of the research community in the use of NNs for surrogate modelling applied to design optimization. The article discusses the effect in optimization of the type of NN selected, the type of data used for the training and if special data treatment is needed, and it reveals the potential of using NNs for optimization purposes. For instance, the authors show that multi-level surrogate modelling, where various levels of information are used, is highly effective to obtain the optimal design area with a small computational cost. NNs can be combined with several optimization algorithms and can also make use of large databases, extracting relevant information suitable for optimization purposes. In this review paper, the authors reveal the relevance of deep learning in aerodynamic design, which reduces the manual interaction in the optimization process.  In the line of using NNs for design optimization, it is worth to mention the work carried out by the following authors during the last 10 years. Chen {\em et al.}~\cite{Chenetal2010} used NNs, used a supervised algorithm combining Self-Organizing Map networks, for selecting referenced airfoils, with Back-Propagaion, to learn the relationship between the aerodynamic performance and airfoil geometry, to estimate airfoil aerodynamic characteristics for aerodynamic design. Sun {\em et al}~\cite{Sunetal2015} introduced an inverse design method using NNs in trained in wing and airfoil databases. Koziel \& Leifsson~\cite{KozielLeifsson2016} developed a methodology for solving transonic airfoil aerodynamic shape optimization problems. The authors combined a low-fidelity CFD model (coarse mesh and relaxed convergence criteria) with NNs, in charge of correcting the low-fidelity simulations with space mapping. The method was able to maximize lift and minimize drag in a two-dimensional transonic flow. Oktay {\em et al.}~\cite{Oktayetal2017} combined NNs with the algorithm of simultaneous perturbation stochastic approximation (SPSA) to estimate the drag coefficient optimum value of a fuselage. By this combination, the authors used SPSA without any objective function, supported by the NNs capabilities, which were trained using wind tunnel data. The NN was generalized to estimate the drag coefficient of different geometry fuselages (different fuselage shape parameters).  More recently, Xu {\em et al.}~\cite{Xuetal2018} used NNs to develop surrogate models, later employed in a genetic algorithm to select optimized airfoil designs for transonic supercritical conditions, with the idea of minimizing time-averaged drag and reducing buffet effect (to prevent structural damages). 

The identification of aerodynamic forces, the estimation of linear and non-linear responses in the aircraft (aeroelasticity and FSI) and design optimization, are three main topics where researchers have put effort since the past to advance in the field of aerospace engineering and aircraft optimal design and manufacturing. Using ML in the field, allows to develop accurate ROMs and surrogate models, suitable for a wide range of problems, from incompressible to transonic compressible flows, at various flight conditions. Nevertheless, the lack of generalization in the methodologies proposed, as well as the possibility of merging different types of data (numerical and experimental), with different types of fidelities, encourage researchers to continue advancing in the field, combining ML with data fusion techniques that will also consider the physical insight of the problem under study. Developing highly accurate predictive model based on machine learning, which will be applicable to solve different problems and requiring small manual interaction, will definitely advance the field of aerospace engineering and the achievement of sustainable aviation.

\section{Aeroacoustics\label{sec:Acourtics}}

The recent review by Bianco et al.  \citep{ doi:10.1121/1.5133944} summarises the advantages of using ML for modelling pure acoustic phenomena. In the review, the authors cover a variety of topics including source localization in speech processing, source localization in ocean acoustics, bioacoustics and environmental sounds. The main ideas in this review include localising noise sources  and identifying acoustic features or patterns using ML. The references included in Ref.~\citep{ doi:10.1121/1.5133944} show that ML-based methods can provide enhanced performance when compared with conventional signal-processing methods. However, ML-based techniques are limited by the existing data, since large amounts of data are typically necessary for training.

More relevant to aeronautical applications is the field of aeroacoustics and computational aero-acoustics (CAA). Aeroacoustics problems are generally more complex than pure acoustics once since the coupling between the turbulent flow interacting with the geometries, the associated acoustic generation and subsequent propagation need to be considered. The most important sources of noise in aeronautical applications include aerodynamic noise, jet noise and rotating machinery noise, see Moreau \cite{Moreau} for the more recent advances on these fields. To the authors’ knowledge there is no review covering the use of ML to model aeroacoustics or to accelerate CAA simulations. 

Aeroacoustics and CAA exhibit two distinct problems: noise generation (near field) and noise propagation (far field). Although direct simulation of acoustics (near and far fields are solved together) is possible, and only for small cases to date, the hybrid approach is generally preferred to lower the cost in large complex simulations. Acoustic analogies revolutionised the field of aeroacoustics in the 1960s (see Lighthill \cite{doi:10.1098/rspa.1952.0060,doi:10.1098/rspa.1954.0049}), and have enabled a certain degree of decoupling between the generation and propagation of noise. Following these analogies, equivalent acoustic sources can be extracted from the well-resolved near-field region. These can be subsequently modelled and inputted into a new simulation to predict their propagation. Simulations of aero-acoustics can use this decoupling to enable faster simulations, since the near field that requires highly-accurate and costly flow computations can be performed separately from the acoustic propagation (wider region but coarser mesh). The field has evolved considerably due to the development of faster computers that enable larger computations, see Lele \& Nichols \cite{Lele} or Moreau \cite{Moreau} . 

Following the acoustic analogies, models or correlations can be applied to either the generation or the propagation regions, or even to both regions at the same time. The model (or correlation, regression) can use NNs. 
%
%
When considering only propagation within the aeroacoustic context, we find only a few works with interesting ideas. For example, Alguacil et al. \cite{ALGUACIL2021116285} predicted the propagation of acoustic waves using deep CNNs and developed deep-learning surrogates to characterises the transfer function for the propagation and scattering of acoustic waves in quiescent flows \cite{doi:10.2514/6.2021-2239}. 
Tang et al. \cite{9417734}  studied dynamic interactive sound propagation by means of neural-network-based-learned scattered fields coupled with ray tracing to generate specular, diffuse, diffraction, and occlusion effects.
Kužnar et al. \cite{KUZNAR20121053} proposed to use linear-regression models to improve vehicle acoustics.

 When considering generation and propagation of aeroacoustics, artificial neural networks are being deployed to model/correlate a variety of input parameters to sound pressure levels. The use of ML in acoustics is scarce and has been restricted to the automotive industry or simple flows. 

Beigmoradi et al. \cite{BEIGMORADI2014123} used Taguchi methods and NNs to reduce the computational complexity for the aeroacoustics of a rear end simplified car model. The model was used for optimisation using genetic algorithms.
R{\"u}ttgers et al. \cite{10.1007/978-3-030-59851-8_6} trained a deep NN using a Lattice-Botlzman method to predict the aerocoustic far field of a 2D square domain that include randomly distributed rectangular and circular objects, and monopole sources.
Moeen Uddin et al. \cite{DOI1} used acoustic analogies based on RANS and Ffowcs Williams and Hawkings, to predict the aero-acoustic tire noise at near-field and far-field receivers around the tire. Artificial neural networks-based regression were used to study the highly non-linear relationships in the system:  between A-weighted sound pressure level and tire parameters (Groove depth, Groove width, Temperature and velocity). Abreu {\it et al.}~\cite{abreu_tanarro_cavalieri_schlatter_vinuesa_hanifi_henningson_2021} have used spectral POD to characterize the wave packets originated at the trailing edge of turbulent airfoils. Note that these structures have important implications in aircraft noise generation, and ROMs aimed at predicting their behavior can help to design more-silent aerodynamic solutions. Finally, Kou {\it et al.}~\cite{Kou_ferrer} used autoencoders to improve the the optimization of airfoils where a multidisciplinary objective function combined aerodynamic and aero-acoustic targets.

The main conclusions of this section is that there is huge potential in using ML methods for aeroacoustics. This field is at its infancy and has not yet been deployed to solve aeronautical applications. The main difficulty is the that the aeroacoustics transfer functions (linking flow/geometrical parameters and far field acoustics) are not always smooth and care is needed when performing tasks involving derivatives or such functions ({\it e.g.,} gradient or adjoint based optimisation). NNs do not always perform well when modelling non-smooth functions \cite{Imaizumi2019DeepNN}, and therefore their use in this field needs caution. A possible way to alleviate the lack of smoothness is to decouple noise generation and  propagation and use NNs separately for each part. Alternatively one may regularise the transfer function using for example smooth flow models ({\it e.g.} based on RANS equations).

\section{Combustion\label{sec:combustion}}

The ultimate goal of combustion research is to develop accurate, generalizable and predictive models to describe the phenomena occurring in combustion devices, which primarily involve turbulent flows. The recent years have witnessed significant advances in the fields of DNS and LES, as well as in the development and reduction of chemical kinetics mechanisms for hydrocarbon fuels. The currently available computing power allows to perform parametric DNS studies closer to the conditions of practical interest. LES is becoming more and more present in industry, although RANS remains the preferred option for its affordable computation cost. Despite the significant progress, existing modelling approaches generally lack generality and predictivity \cite{Brunton2020}. The main challenges are associated with the number of species involved in combustion processes, the small scales and the non-linear turbulence-chemistry interactions \cite{Bilger2011, Pope2013, Giusti2019} ML methods can contribute to combustion science and treat some of the previously unmet challenges, providing interpretable feature extraction techniques, delivering generally applicable approaches to locally adapt comprehensive chemical mechanisms and sub-grid models, designing new closures to parametrise the unresolved fluctuations, and developing ROMs for fast and yet-accurate system evaluations. 
In the following, the current state and perspective associated to the use of ML in combustion physics are reviewed and discussed.

\subsection{Data analysis and feature extraction\label{sec:AnalysisFeature}}

Combustion science has traditionally dealt with massive amounts of data from experiments and large-scale numerical simulations. As a matter of fact, big data has been a reality in combustion for almost three decades \cite{Chen2011} thanks to the advent of petascale computing and, hence, the possibility of performing fine-grained simulations of canonical and laboratory-scale turbulent flames with detailed chemistry. While exascale computing will allow to investigate closer-to-reality conditions, DNS of turbulent combustion already represents a key research area for model development and validation \cite{Trisjono2015, Wick2020, pitsch2020data}. Over the past 50 years, many techniques have been developed to handle and process combustion data from experiments and simulations. While these analyses have strongly relied on domain expertise and heuristic algorithms, they can be certainly regarded as early applications of ML.
Several early studies focused on the use of dimensionality reduction techniques such as PCA \cite{Jolliffe1986}, with the objective of identifying empirical low-dimensional manifolds in combustion systems \cite{Maas1998, Frouzakis2000, Danby2006, Parente2009, Sutherland2009, Parente2011, Biglari2012, Parente2013, Mirgolbabaei2013, Yang2013}. The analysis focused both on experimental \cite{Barlow1998, Barlow2000, Barlow2005, DALLY20021147, Schneider2003, Medwell2007} and numerical simulation data \cite{Echekki2003, Hawkes2007, Hawkes2009, Yoo2011, Punati2011, Coussement2012, Coussement2013, Bansal2015}. The increasing availability of DNS data of complex configurations is shifting the interest towards computational data, for the possibility of mining large sets and accessing variables not available from experiments ({\it e.g.,} source terms and turbulence-related variables). The dimensionality reduction offered by PCA can be limited for combustion data. Indeed, PCA is a multi-linear technique and typical reacting flow data exhibit strong non-linearities. This implies that additional components are identified by PCA to make up for the data non-linearity, leading to an overestimation of the true problem dimensionality.
To overcome the limitations of classic PCA with non-linear systems, the application of non-linear methods has been proposed in the community, including non-linear and KPCA \cite{Mirgolbabaei2013, Coussement2012, MIRGOLBABAEI20144622}, isometric mapping (IsoMap) \cite{Bansal2011}, T-distributed Stochastic Neighbor Embedding (T-SNE) \cite{Fooladgar2018, Fooladgar2019} and AEs \cite{zdybal2022advancing, GANGOPADHYAY2021100067}. Recently, data-driven approaches based on CNNs were proposed for regime identification \cite{Zhu2019, Wan2020}. Moreover, non-linear regression based on Neural Networks \cite{Jigjid2021} was employed to identify the main features of Moderate or Intense Low-oxygen combustion (MILD)\footnote{MILD combustion is a combustion technology able to ensure very high combustion efficiencies with no soot, and very low nitrogen oxides and noise emissions, compared to conventional combustion regimes, due to the reduced temperature peaks and macro-scale homogeneity. In addition, it is a fuel-flexible technology. } and predict the co-existence of different combustion modes, paving the way to the adaptive selection of closures during numerical simulations. Non-linear methods greatly improve the dimensionality reduction potential. However, they become computationally intractable for large data sets (KPCA and IsoMap) or involve many hyper-parameters to optimise and cross validate (t-SNE, NNs and CNNs). Finally, non-linear algorithms extract features that can be more challenging to interpret and that cannot be easily converted into predictive modelling approaches.
An alternative to non-linear methods is to identify local principal curves \cite{Einbeck2005} and locally-linear embeddings (LLE) \cite{Roweis2000}, as well as to perform local modal decomposition, {\it i.e.,} applying PCA locally (LPCA) \cite{KambhatlaL97}. Algorithms based on the definition of an {\it a priori} prescribed (supervised) conditioning variable (such as mixture fraction), and on the iterative (unsupervised) minimization of the reconstruction error have been investigated in the context of combustion data analysis \cite{Parente2011}. The latter has proven particularly effective and even competitive with more sophisticated approaches such as AEs for dimensionality reduction \cite{zdybal2022advancing}, as well as for clustering tasks \cite{DALESSIO202068, DAlessio2020EN}. Coupling vector quantization with dimensionality reduction using PCA (VQPCA) \cite{Parente2011, KambhatlaL97} provide robust classification algorithms, less prone to overfitting and applicable to conditions different than those met during training \cite{Dalessio2021}. Indeed, VQPCA requires the determination of only two hyper-parameters, the number of eigenvalues, controlled by the amount of desired variance, and clusters, based on quantitative metrics \cite{Fu1977, Calinski, Rousseeuw, Tibshirani} as well as the statistical representativeness of the clusters \cite{DALESSIO202068}. The classification properties of LPCA have been exploited to identify the leading principal variables (PV) and processes in different regions of the state-space \cite{Li2021,PeppeBook, PeppeSOCO, Valorani2020BOOK}, and to develop adaptive chemistry models \cite{DALESSIO202068, DAlessio2020EN}. 

\subsection{Dimensionality reduction, classification and adaptive chemistry\label{sec:Dimensionalityred}}

The large dimensionality of combustion systems has driven the development of dimensionality reduction approaches. Considering the impact of large kinetic mechanisms on the computational time of detailed numerical simulations \cite{Pope2013}, many different strategies have been developed in the last two decades for the systematic generation of reduced mechanisms, including Sensitivity Analysis \cite{Turanyi1990}, PCA \cite{Turanyi1990, Griffiths1995}, Path Flux analysis \cite{Frouzakis2000}, graph-search methods such as Directed Relation Graph (DRG) \cite{Lu2005}, DRG-aided sensitivity analysis \cite{Sankaran2007}, DRG with error propagation (DRGEP) \cite{PEPIOTDESJARDINS200867}, lumping \cite{Huang2005}, quasi-steady state approximation (QSS) using Computational Singular Perturbation (CSP) \cite{Valorani2006}, Tangential Stretch Rate \cite{MALPICAGALASSI2018439}, Level of Importance (LOI) \cite{LOVS20001809}, via production/consumption analysis \cite{Zambon2007}, and using error estimation \cite{TOMLIN1997293}. An effective strategy to reduce the burden of detailed chemistry is to use adaptive chemistry approaches, which adjust the mechanisms to the local flow conditions \cite{Liang2009, Contino2011, Zhou2016, Ren2014, Ren2014DAC, Shi2010}, performing the reduction on the fly. Approaches relying on pre-tabulated libraries of reduced mechanisms \cite{DALESSIO202068, DAlessio2020EN, Dalessio2021, Liang2015, Newale2019, Newale2021} have been recently proposed, thanks to the development of efficient classification algorithms. This opens the way to the use of reduction methods characterised by higher overhead and unsuited for use at runtime, {\it i.e.,} error-controlled approaches \cite{PEPIOTDESJARDINS200867, Stagni2016, Niemeyer2010} and methods relying on the dynamical description of the system \cite{Valorani2006, MALPICAGALASSI2018439}.

In order to cope with the large number of uncertain parameters in comprehensive chemical mechanisms, uncertainty quantification and optimization have been adopted in the process of chemical mechanism development \cite{Frenklach2007, Tomlin2013, WANG20151, MILLER2021100886}. In particular, several methodologies have been designed to solve the so called “inverse problem” \cite{FRENKLACH199247, SHEEN20112358}, and improve predictions \cite{VARGA2015589} based on increasingly available experimental data. Notably, the use of genetic algorithms \cite{ELLIOTT2004297, Bertolino2021} has gained momentum to deal with the highly structured objective functions, typical of mechanism optimization, often characterised by the presence of multiple local minima/maxima \cite{FRENKLACH199247}. Considering the increasing amount of kinetic parameters from theoretical {\it ab initio} calculations \cite{MILLER2021100886, Curran2019}, one challenge in optimization approaches is the estimation and/or determination of plausible uncertainty ranges based on the level of theory adopted throughout the calculation protocols for electronic structures, potential energy surfaces and phenomenological reaction rate constants \cite{Bertolino2021, Shannon2015, Klippenstein2017}. The availability of highly-efficient optimization algorithms and tools \cite{FURST2021107940} is pushing research in the development of highly reduced chemical mechanisms for complex fuels, to allow high-fidelity simulations with realistic chemistry rather than global mechanisms \cite{Jaouen2017, JAOUEN201760}. In this line of research, new frameworks combining dimensionality reduction ({\it e.g.,} with PCA) and non-linear regression ({\it e.g.,} using b-spline interpolants and NNs) have been proposed to replace the tabulation of reaction rates in flamelet/progress variable \cite{Christo1996, Blasco1998, Ihme2009, Bode2019} as well as in finite-rate chemistry approaches \cite{Chatzopoulos2013, FRANKE2017245}, and to simplify the chemistry of complex hydrocarbons combining data-based models for representative pyrolysis species and foundational chemical mechanisms for the remaining ones \cite{ALQAHTANI2021142}. Along the same line, novel hybrid frameworks relying on CNNs \cite{Seltz2021} have been proposed to model the dynamics of particles and their size distributions, with application to soot predictions. Finally, virtual chemistry approaches consisting of virtual species and reactions, and employing optimization algorithms have been recently proposed \cite{Cailler2020} to predict quantities of interest ({\it e.g.,} pollutant emissions) in large-scale simulations at a reduced computational cost.
Besides rate-based methods that identify and eliminate redundant species and reactions, approaches based on the re-parameterization of the chemical state-space \cite{PETERS19881231, OIJEN, Gicquel2000, Fiorina2003, Pierce2004} have been further developed and combined to data-driven approaches. In particular, PCA was used to develop reduced-order models for combustion simulations based on the resolution of transport equations for the principal components \cite{Sutherland2009}. To reduce the number of transported components and improve the reconstruction of the state-space as well as the associated chemical source terms, PCA was coupled to non-linear regression techniques, replacing the (multi-linear) PCA mapping between the reduced PC space and the original one. Different regressions have been used to this purpose including Multi Adaptive Regression Splines (MARS) \cite{Biglari2012, Yang2013}, NNs \cite{Mirgolbabaei2013, Dalakoti2020} and Gaussian Process Regression (GPR) \cite{Isaac2015}. This modelling approach has been demonstrated for simple reactors \cite{Mirgolbabaei2013, Isaac2015, Malik2018} One Dimensional Turbulence (ODT) simulations \cite{Mirgolbabaei2013, Biglari2015}, premixed and non-premixed DNS simulations \cite{Coussement2012, Coussement2013, Coussement2016, Owoyele2017}, as well as in the context of RANS simulations using kernel density estimation (KDE) \cite{Ranade2019, RANADE2019279}. Recently, the method combining PCA and GPR (PC-GPR) was applied, for the first time, to the three-dimensional LES simulation of flames D-F \cite{Malik2021} and of the Cabra flame \cite{MALIK2022112134}, showing excellent predictive capabilities. The ability of the method to generalise the formulation of flamelet/progress variables approaches and facilitate the formulation of a sub-grid closures (being the selected scalars uncorrelated) appears particularly promising.

\subsection{Combustion closures\label{sec:closures}}

In turbulent flows, temperature fluctuations can be as high as several hundred Kelvins. Considering that Arrhenius reaction rates are highly non-linear functions of temperature, accurate statistical closures for filtered approaches cannot be based on an expansion about mean properties \cite{Pope2013, poinsot:hal-00270731}. The task of a combustion model is to provide a description of the unresolved scales based on the information available during a simulation. Combustion models have been often classified into two categories, the flamelet-like \footnote{Flamelet-like models  are based on the re-parametrisation of the thermo-chemical state using a reduced number of variables identifying a low-dimensional manifold in the composition space on which the evolution of the system is constrained} \cite{PETERS19881231, OIJEN, Gicquel2000, Fiorina2003, Pierce2004} and PDF-like approaches \footnote{PDF-like approaches do not make the assumption of a low-dimensional manifolds and treat the mean/filtered reaction rates exactly, solving the, one-point one-time, joint PDF of fluid composition. Mixing at the molecular level requires a closure in PDF approaches.} \cite{pope_2000, fox_2003, Haworth2011}. Besides them, we can also mention reactor-based \footnote{Reactor-based models assumes that a fraction or the entire computational cell behaves like an idealised reactor (perfectly-stirred reactor or plug-flow reactor), allowing an affordable treatment of finite-rate chemistry.} \cite{MAGNUSSEN1977719, Chomiak, Ferrarotti2019, Li2019} and conditional-moment approaches (CMC) \footnote{CMC relies on the concept of conditional manifolds and equations are derived and modeled for the conditional moments of the reactive scalars.} \cite{KLIMENKO1999595}. Data-driven methods are significantly impacting the design of improved sub-grid models for LES and RANS simulations, beside turbulence closures \cite{Bode2021, Liu2020}. Recently, Deep CNNs have been used by different authors, to improve the prediction of the unresolved flame surface wrinkling \cite{lapeyre_et_al} in the context of the Thickened Flame model \cite{Colin2000}, as well as to predict the filtered progress variable source term and the unresolved scalar transport terms in LES \cite{Nikolaou2019, Seltz2019, Nikolaou2021}. NNs and Deep Learning (DL) were employed to develop presumed probability-density-function (PDF) models for conventional \cite{THENRYDEFRAHAN2019436} and MILD combustion \cite{Chen2021}. Data-driven approaches were recently used to assess the validity of existing reactor-based closures in MILD combustion \cite{Iavarone2021, PequinPoF}, to obtain conditional statistics \cite{Yao2020, Yao2021} in the context of CMC, and for the dynamic assignment of combustion closures \cite{Chung2021}.

\subsection{Reduced-order models for realistic combustion systems\label{sec:roms}}
In some applications, a very fast evaluation of a system response is required (for control, optimization, …), thus limiting the use of time-consuming simulations. In this context, ROMs are used to approximate the underlying hidden relationship between inputs and outputs, using available observations to estimate the system response for unexplored conditions. Black-box approaches have been widely used in combustion to create static input-output maps \cite{PALME20113898} and for system identification \cite{KALOGIROU2003515}, to predict macroscopic quantities such as exhaust gas emissions and temperature and to detect oscillatory patterns such as thermoacoustic instabilities, respectively. In combustion, NNs have been used to evaluate and tabulate reaction rates as discussed above \cite{Christo1996, Blasco1998, Ihme2009, Bode2019}, as well as to estimate modelling errors in LES \cite{Trisjono2015, Berger2018}. Black-box approaches can be very powerful although not equipped with the guarantees of physics-based approaches. To cope with that, nonlinear system identification techniques were proposed to enlighten black box approaches by discovering the underlying physics \cite{Brunton2016, Champion2019}. 
An interesting approach to generate combustion ROMs is the use of reactor network models \cite{SWITHENBANK1973627, Falcitelli2002, Lyra2013}, which fall in the category of grey-box models, as they combine a theoretical structure, {\it i.e.,} the canonical reactors, with data to generate the network. The approach can be also regarded as an example of multi-fidelity methods, relying on high-fidelity tools ({\it i.e.,} CFD) to construct the reactor network and the lower-fidelity network model to evaluate pollutant emissions and other quantities, using detailed chemical mechanisms. These techniques are particularly useful for large and complex systems, such as furnaces or gas turbines. However, their current overall fidelity and generalizability are limited by the high-fidelity simulations required to generate the network structure, indicating the interest for data-driven approaches in this area of research. 
ROMs based on projection methods (such as PCA and other such techniques) have been employed for experimental and numerical combustion data \cite{Maas1998, Frouzakis2000, Danby2006, Parente2009, Sutherland2009, Parente2011, Biglari2012, Parente2013, Mirgolbabaei2013, Yang2013}. The direct evolution of the modal coefficients in CFD codes ({\it i.e.,} intrusive ROMs) has been quite limited for non-linear problems like combustion, due to the high cost of the associated Galerkin projection as well as the complexity of coupling them to legacy and commercial CFD codes \cite{McQuarrie2021}. The application of non-intrusive projection-based ROMs has been, on the other hand, quite successful, in combination with non-linear regression approaches such as GPR \cite{Aversano2019PCAKriging}, NNs and Polynomial Chaos Expansion (PCE) \cite{Aversano2021PCEKr}. Indeed, when the number of variables of interest is high, many ROMs need to be trained. Besides, any correlation between these variables is lost in the process of training individual ROMs. Introducing PCA and similar techniques, the number of ROMs can be reduced as the original variables are encoded into a set of fewer uncorrelated scalars explaining most of the system variance. In the combustion community, the combination of dimensionality reduction (using PCA) and non-linear regression (using GPR) was used to develop ROMs for uncertainty quantification \cite{Aversano2019} as well as to design simulation-based digital twins  \cite{Aversano2021} of industrial systems. Indeed, combining CFD simulations with real-time data coming from sensors is possible only if the prediction of a new system state, based on continuous incoming data, becomes instantaneous \cite{Aversano2021, Kraft2017}.

\section{Structural assessment\label{sec:Solid}}

Aerospace structures are subjected to a wide spectrum of loads, operating in harsh environments \cite{Ross2016} and phenomena such as high-velocity debris have the ability to severely impact the integrity of components \cite{Rocha2021}. Furthermore, while developments in composite materials present large advantages in terms of strength and  weight, failure mechanisms are more difficult to predict as a result of an increased number of failure modes compared to metals \cite{Rocha2021}.

Structural health monitoring (SHM) is the implementation of a damage-detection strategy for engineering structures \cite{WordenFarrarSHM}. Applied across many industries, the integration of SHM into the aerospace sector is particularly beneficial as a result of the difficulty of regular inspection and the very high cost of failure. Integrating SHM into aerospace structures has the capacity to reduce the downtime, operating and maintenance costs of aircraft, while also improving the safety and reliability simultaneously \cite{Yuan2016}. 

While vibration-based techniques are the most commonly employed \cite{WordenFarrarSHM}, acoustic emission and electro-mechanical impedance (EMI) are examples of sensing techniques that are often discussed in the context of damage detection within SHM for aerospace structures.  The former uses sensors to detect high-frequency stress waves that are generated during crack propagation and, while aerospace applications have been considered since the 1970s \cite{1976Harris}, significantly enhanced accuracy of crack-location prediction has been achieved for aerospace applications using ML methods such as Gaussian process regression \cite{Hensman2010211} and NNs \cite{Bhat2003} in recent years. Further examples of the use of ML methods being used for damage detection classification within the aerospace industry include probabilistic neural networks \cite{Giurgiutiu2005}, support-vector machines \cite{Loutas2012} and logistic regression \cite{Jiang2021}.

One problem facing damage detection using SHM methods is the requirement of complex and heavy wiring requirements of the sensor networks which can be prohibitive for aerospace structures \cite{Gianni2020}. As a result, \textit{low power} \cite{Gianni2020} or \textit{no power} equipment \cite{Rubes2021} have been developed to enable wireless sensor networks. Such approaches require efficient methods of communicating data and, thus, new methods of interpreting the data have been developed. In Ref. \cite{Salehi2018}, low-rank-approximation is used to estimate sparse data from a wireless sensor network and a K-nearest-neighbour algorithm is employed to classify this imputed data. Furthermore, sensor and wiring reliability can be considered to be one of the ‘weak-links’ of an effective SHM system, particularly in the harsh environments aerospace structures are subjected to, and classification techniques like KNN can be used to detect sensors faults \cite{Melia2016}.

Digital twins are widely discussed across a broad range of engineering domains, where in SHM applications, they are developed with an aim of providing an accessible insight into the health of a structure or inform predictive maintenance. This is achieved by having a digital model of the structure in which measured detail can be incorporated \cite{Tuegel2011}.  A common example of this within the aerospace sector is performing updated finite-element modelling; however, this can be computationally costly. ML can, therefore, be used to perform surrogate modelling using methods such as Gaussian processes \cite{Li2017b, Chakraborty2021} to significantly reduce the computational cost.  

Many aerospace components such as landing gear operate a safe-life methodology, in which there is zero-crack tolerance \cite{Hoole2016}. Therefore, damage-detection methods, based {\it e.g.} on Acoustic emission, are not always relevant (although their inclusion is advocated, nonetheless). A sub-category of SHM is \textit{structural loads monitoring} or \textit{virtual sensing}. This involves predicting the loads, as well as the subsequent stress and strain, that a structure is subjected to using incomplete available measurements. This enables damage prognosis, which can be considered the natural next step of structural health monitoring with applications in all forms of defence and commercial aviation \cite{Farrar2017}.  Fatigue analysis is of particular interest here, as more than 80\% of structures fail due to fatigue \cite{wirsching1995probabilistic}, which occurs after the structure is subjected to a large number of stress cycles. 

Within the helicopter health and usage monitoring systems (HUMS) domain, there are numerous ML-based approaches to loads monitoring and fatigue analysis. In Ref. \cite{Manry1999}, the mechanical loads are predicted using a multilayer perceptron (MLP) NN. In Ref. \cite{Valdes2017}, an extreme learning machine (a subclass of feed-forward NN)  is used for load prediction and the results are fed through the \textit{stress-life} fatigue analysis procedure to predict damage accumulation. 

\textit{Grey-box modelling} is the integration of our knowledge of the physics of the structure into the model. The most common way of doing this, and something the authors advocate for whenever possible, is via \textit{semi-physical} modelling or \textit{input augmentation}, involving manipulating inputs to the model to indicate physical processes \cite{reed2005parametric, fuentes2014aircraft, Azzam1997}. An example of this is squaring the measured airspeed before using it as an input to the model (in this case a NN) to give an indication of dynamic pressure \cite{reed2005parametric,reed2007development}. Similarly, in Ref. \cite{Azzam1997}, a \textit{mathematical network} is used for fatigue monitoring, in which physics-inspired merging functions are used as the first layer to the network. An alternative way of integrating physics into models is via \textit{residual modelling} in which ML methods such as Gaussian process regression \cite{pitchforth2021grey,wan2015residual} is used to predict the discrepancy between a physics-based or empirical prediction and the measured data. An introduction to grey-box modelling for SHM applications can be found in Ref. \cite{cross2022physics}.  

A data-driven model is fundamentally an estimate of the similarity of the test case compared to the training data that the model has seen. Practically, it is difficult to include all possible loading phenomena of aerospace structures in the training dataset due to magnitude of the loading spectrum and computational limits. Therefore, probabilistic methods are becoming increasingly popular. Gaussian process regression is one of the most popular ML methods for virtual sensing problems, in part, due to its inherently probabilistic nature and subsequent capacity to indicate uncertainty. In the case of Gaussian process regression, the estimate of the similarity of the training and the test data is based on the \textit{kernel}, or \textit{covariance function}. Based on prior knowledge of the data, an appropriate kernel can be chosen and combining kernels using addition and multiplication enables a rich language that the functional form of the model can take \cite{duvenaud2014automatic}. A novel way of inputting physical insight into Gaussian process regression models is to integrate the knowledge into the kernel itself, which has the added benefit of providing a physical interpretation to the samples drawn from the model posterior \cite{cross2022physics}.

One example of Gaussian process regression used for virtual loads monitoring can be found in Ref. \cite{holmes2016prediction}, where the loads on the landing gear of an aircraft are predicted with a high degree of accuracy. In Ref. \cite{fuentes2014aircraft}, the loads on an aircraft wing are predicted and fatigue analysis is carried out using the predicted strain values.  However, the probabilistic aspect of the model is often not fully utilised. In Ref. \cite{Gibson2020}, Gaussian process regression is again used to predict the stress on the wing of an aircraft. By taking a large number of draws from the posterior covariance of the model, the model uncertainty can be propagated through the stress-life fatigue analysis procedure and, thereby, a probabilistic estimate for fatigue damage accumulation is developed. 

Such approaches can fit into probabilistic risk-based decision frameworks for structural health monitoring \cite{hughes2021probabilistic}, which can be used to make risk-informed decisions for safety-critical components like those found in the aerospace industry. Adopting a probabilistic approach to fatigue in aerospace components has the potential to better control the risk level and plan maintenance more effectively \cite{cavallini2007probabilistic,Hoole2021}.

\section{Conclusions\label{sec:conclusions}}

The continued growth of machine learning (ML) is leading to a progressively larger impact on a wide range of scientific areas, including aerospace engineering. A number of emerging ML-based technologies are already impacting every aspect of this area, e.g. in terms of simulation capabilities and enhanced physical insight. The main goal pursued by the aerospace industry is to use ML to develop several applications connected to the reduction of aircraft’s environmental impact, also considering system management, customer service, data interpretation and even developing new methods based on ML-tools capable to generate new high-fidelity databases at a reduced cost, which will be later used to develop new designs, more efficient, in a fast and efficient way. These applications involve acquiring knowledge and developing new ML tools suitable to solve problems of fundamental fluid dynamics (experimental and numerical), aerodynamics, acoustics, combustion and structural health monitoring. 

The type of identification method, the type of data analysed, and the model structure, influences the performance of the ML model. To maintain the proper balance between the model accuracy and the cost of data generation, a suitable option would be using ML tools considering data fusion techniques, which combine data from different types (experimental and numerical) and different fidelities. The development of models that are generalizable within the different applications, would suppose and advance in the field, reducing in this way the complex process of calibration behind selecting the proper architecture composing the neural networks composing the model
and extracting relevant information from the different data sources available, which is highly dependant on the type of application:  aerodynamics, combustion, acoustics, solid mechanics, etc. However, most of the models currently used by the industry, only consider data coming from a single source, leaving new research opportunities to advance in the field. Also, considering models grounded in physics, which combine modern neural network architectures with other dimensionality reduction techniques based on the identification of physical patterns, would suppose and advance in the field. Some researchers are currently developing their work following these ideas, showing the good performance of these models grounded in physics: a few examples are found in the fields of fundamental fluid mechanics, aerodynamics combustion and structural health monitoring, nevertheless there are still many open questions that should be address, and also the lack of robust and generalizable schemes still leaves it as an open topic that should be address by researchers and the aerospace industry in the near future.

Finally, when it comes to the application of ML to CFD, there is great potential in several areas related to improving the efficiency of numerical simulations of fluid flows. The future development of ML should not be focused on replacing CFD, but rather improving the efficiency and robustness of subproblems within CFD solvers. For instance, regarding acceleration of CFD ({\it e.g.,} by accelerating the Poisson problem or improving coarse simulations), enhancing modeling (for LES and RANS) and improving ROM development. In the latter, AEs are progressively becoming a more useful technique since they enable leveraging the non-linearity while, through recent developments~\cite{ae_modal}, retaining the orthogonality of the AE modes, which are ranked by their contribution to the reconstruction as in POD. One point of future improvement for AEs is the possibility to effectively integrate the temporal dynamics into the predictions, a step that currently relies on using the decoder and the latent vectors. When it comes to wind-tunnel experiments, non-intrusive sensing has greatly benefited from recent progress in deep-learning architectures for computer vision. Another area holding great potential is that of flow control, where a number of methods ({\it e.g.,} Gaussian-process regression, genetic programming and deep reinforcement learning) have the potential of discovering very effective strategies.

\section*{Acknowledgements}
E.F. and S.L.C. would like to thank the support of the the Comunidad de Madrid through the call Research Grants for Young Investigators from the Universidad Politécnica de Madrid. E.F. also acknowledges the Spanish Ministry MCIN/AEI/10.13039/501100011033 and the European Union NextGenerationEU/PRTR for the grant ``Europa Investigación 2020'' EIN2020-112255. S.L.C. also acknowledges the grant PID2020-114173RB-I00 funded by MCIN/AEI/10.13039/501100011033. 

\small
\bibliographystyle{abbrvnat}
\bibliography{mlaero_bib}
 
\end{document}